%% file: main.tex
\pdfoutput=1
\def\arxivversion{}
\documentclass[10pt,twocolumn,letterpaper]{article}

\ifdefined\arxivversion
\usepackage[pagenumbers]{cvpr}
\makeatletter
\let\cvpr@maketitle\@maketitle
\def\arxiv@maketitle{%
  \newpage\null\vskip.375in\relax
  \begin{center}
    {\Large\bf\@title\par}
    \vspace*{\arxiv@above}
    {\large\lineskip.5em\begin{tabular}[t]{c}\@author\end{tabular}\par}
    \vspace*{\dimexpr\arxiv@free-\arxiv@above\relax}
  \end{center}}
\def\@maketitle{%
  \setbox\z@\vbox{\def\@author{Anonymous \confName~submission}\cvpr@maketitle}%
  \def\arxiv@free{0pt}\def\arxiv@above{0pt}\setbox\tw@\vbox{\arxiv@maketitle}%
  \edef\arxiv@free{\the\dimexpr\ht\z@-\ht\tw@\relax}%
  \edef\arxiv@above{\the\dimexpr\arxiv@free/2+6pt\relax}%
  \arxiv@maketitle}
\makeatother
\else
\usepackage[review]{cvpr}
\fi

\input{preamble}

\definecolor{cvprblue}{rgb}{0.21,0.49,0.74}

\def\confName{CVPR}
\def\confYear{2027}
\usepackage{url}
\usepackage{booktabs}
\usepackage{amsfonts}
\usepackage{nicefrac}
\usepackage{microtype}
\usepackage{xcolor}
\usepackage{lipsum}

\definecolor{dblue}{RGB}{52, 104, 192}
\definecolor{dgreen}{RGB}{65, 171, 93}
\definecolor{dred}{RGB}{210, 69, 69}
\definecolor{dpurple}{RGB}{148, 103, 189}
\definecolor{dorange}{RGB}{255, 127, 14}
\definecolor{dcyan}{RGB}{35, 70, 100}
\colorlet{figObj}{dblue}
\definecolor{figHOI}{HTML}{A0457A}
\colorlet{figEval}{dorange!92!black}
\definecolor{matRigid}{HTML}{2A78D6}
\definecolor{matDeform}{HTML}{EB6834}
\definecolor{matCodim}{HTML}{1BAF7A}
\definecolor{matFlow}{HTML}{EDA100}

\usepackage{bbm}
\usepackage{graphicx}
\usepackage{amsmath}
\usepackage{amssymb}
\usepackage{amsthm}
\usepackage{multirow}
\usepackage{caption}
\usepackage{wrapfig}
\usepackage{placeins}
\usepackage{enumitem}
\usepackage[colorlinks,
            pagebackref,
            breaklinks,
            linkcolor=red,
            anchorcolor=magenta,
            urlcolor=dblue,
            citecolor=dblue
            ]{hyperref}
\usepackage{subcaption}
\usepackage{multicol}
\usepackage{colortbl}
\usepackage{xspace}
\usepackage{siunitx}
\usepackage{array}
\newcommand{\tpm}[1]{\,{\scriptsize$\pm$#1}}
\newlength{\alnumL}\newlength{\alnumR}
\newcolumntype{N}[2]{>{\settowidth{\alnumL}{$#1$}\settowidth{\alnumR}{\tpm{#2}}}c}
\NewDocumentCommand{\alignednum}{m o}{%
  \makebox[\alnumL][r]{#1}\makebox[\alnumR][l]{\IfValueT{#2}{\textcolor{black!55}{\tpm{#2}}}}}
\usepackage{pifont}
\usepackage{tikz}
\usetikzlibrary{positioning,arrows.meta,fit,calc,backgrounds,shapes.callouts}
\AtEndPreamble{\crefname{supp}{Supp.}{Supp.}\Crefname{supp}{Supp.}{Supp.}}

\input{symbols}
\input{math_commands}

\title{CodeHOI: Code with Agents for Physically Plausible 4D Hand-Object Interaction}

\ifdefined\arxivversion
\author{Seonghwan Park\qquad Inho Chang\qquad Min-Gyu Park\\[4pt]
Korea Electronics Technology Institute (KETI)
}
\fi

\begin{document}
\twocolumn[{%
\renewcommand\twocolumn[1][]{#1}%
\maketitle
\begin{center}
\includegraphics[width=\linewidth]{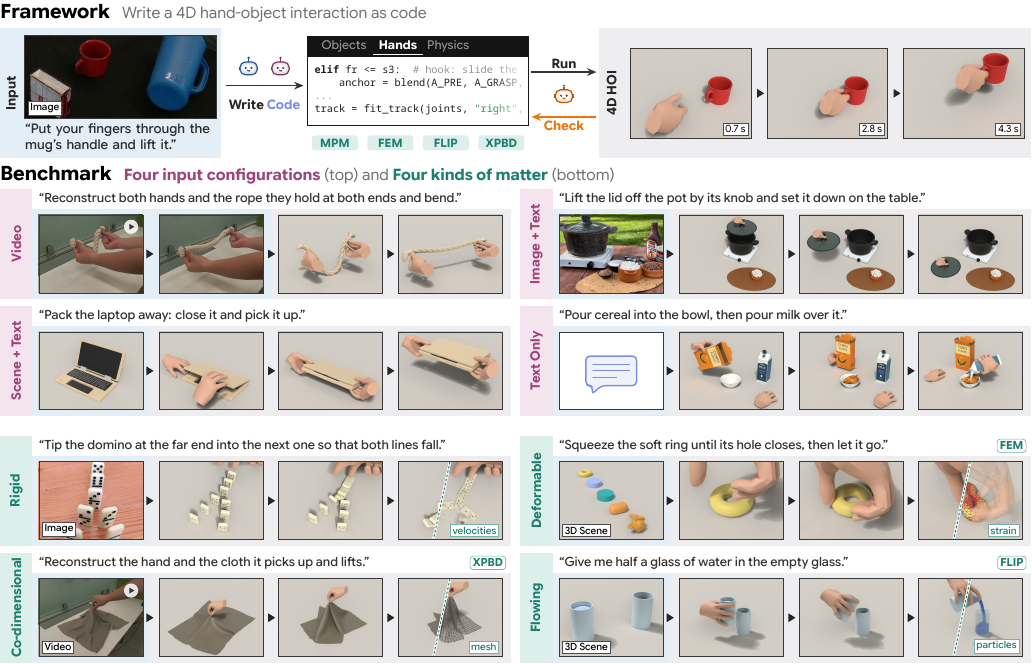}
\captionof{figure}{\method writes physically plausible 4D hand-object interaction (HOI) as a program.
(Top) From an input and an instruction, agents write the objects, hands and physics as code, run it with established physics solvers (MPM, FEM, FLIP or XPBD) and check the resulting 4D HOI against physical laws.
(Bottom) \bench evaluates reconstruction from a video and generation from an image, a 3D scene or text alone, across four kinds of matter. For each kind, the last frame is cut open to reveal the simulated state.}
\label{fig:teaser}
\end{center}%
\vspace{0.4em}
}]
\input{sec/0_Abstract}
\input{sec/1_Introduction}
\input{sec/2_Related_work}
\input{sec/3_Method}
\input{sec/4_Evaluation}
\input{sec/5_Conclusion}

{
    \small
    \bibliographystyle{ieeenat_fullname}
    \bibliography{main}
}

\appendix
\crefalias{section}{supp}\crefalias{subsection}{supp}\crefalias{subsubsection}{supp}
\input{sec/6_Supplementary}

\end{document}

%% file: preamble.tex
\usepackage{longtable}

%% file: symbols.tex
\definecolor{tabgray}{HTML}{f2f2f2}
\definecolor{tablavender}{HTML}{E6F7E6}
\definecolor{tabnavy}{HTML}{00008B}
\definecolor{tabwhite}{HTML}{FFFFFF}
\definecolor{linf}{HTML}{8172b2}
\definecolor{l2}{HTML}{ccb974}
\definecolor{l1}{HTML}{64b5cd}

\definecolor{lavender}{HTML}{9ad756}
\definecolor{DBlue}{RGB}{31, 119, 180}
\definecolor{DGreen}{RGB}{44,160,44}
\definecolor{DRed}{RGB}{214,39,40}
\definecolor{dorange}{RGB}{255, 127, 14}
\definecolor{dorange2}{RGB}{255, 165, 0}

\newcommand{\method}{CodeHOI\xspace}            %
\newcommand{\bench}{CodeHOI-Bench\xspace}       %

\newcommand{\modelicon}[1]{\ifcsname modelicon#1\endcsname\csname modelicon#1\endcsname\fi}

%% file: math_commands.tex
\usepackage{bm, dsfont}

\let\hat\widehat

\makeatletter
\newcommand{\ve}{\@ifnextchar\bgroup{\velong}{{\bm{e}}}}
\newcommand{\velong}[1]{{\bm{#1}}}
\makeatother

%% file: sec/0_Abstract.tex
\begin{abstract}
Modeling 4D hand-object interaction (HOI) is fundamental to understanding manipulation.
This requires physics, since hands move objects only by touch and each material responds differently.
Yet existing methods ignore this physics or simulate only a narrow range of matter.
Programs can instead describe any matter with its physics, but coding agents omit the hand that drives the interaction.
We close this gap with \method, to our knowledge the first agentic framework that writes physically plausible 4D HOI as programs.
To write one, an Object Agent first builds each object with the physics of its matter.
An HOI Agent then specifies the hands by their contacts and simulates each kind of matter with an established physics solver.
Finally, an Evaluator Agent checks each result against physical laws.
We also introduce \bench with 100 reconstruction and generation tasks across four kinds of matter.
With off-the-shelf frontier models, \method writes interactions that are plausible for every kind of matter and remain so when transferred to an independent physics engine.
It thereby enables coding agents to write physically plausible HOI across diverse matter.%
\vspace*{-12pt}%
\end{abstract}

%% file: sec/1_Introduction.tex
\section{Introduction}
\label{sec:intro}

Hands are our primary interface with the physical world, so reconstructing and generating 4D hand-object interaction (HOI) is essential to animation~\cite{zhang2021manipnet}, AR/VR~\cite{han2020megatrack} and robot learning~\cite{qin2022dexmv}.
These applications require HOI that obeys \emph{physics}, where hands act only through contact and each object responds as its material dictates.
Yet data-driven methods ignore this physics and mostly cover rigid objects~\cite{chao2021dexycb,fan2023arctic}, whereas physics-based methods extend to deformable objects~\cite{jiang2025phystwin,lee2026physhandi} driven by tracked hands.
Physically plausible HOI with other matter such as liquids and grains thus remains largely unexplored for both tasks.

Writing the interaction as a program offers a way to close this gap, since a program can build any object, assign its material and simulate how it responds.
Coding agents already write such programs for 3D scenes~\cite{hu2024scenecraft,yang2024holodeck}, articulated objects~\cite{mandi2024real2code,le2025articulate} and dynamic events~\cite{shen2026fourdcodebench,tang2026video2world}, and they verify their results by running the code.
Yet none of them includes the \emph{hand}, whose physical contact with an object determines how the interaction unfolds (\eg, grasp, push or pull).
We therefore ask whether an agent can also write physically plausible HOI as a program.
To do so, the agent must give each object a material that governs its response, make the hands act on objects only through contact under physical laws and verify the result against these laws.

We thus present \method, an agentic framework that writes physically plausible HOI as a program.
It comprises three agents, each addressing one of these requirements:
(i)~\textbf{Object Agent}, which builds the objects in a scene and gives each the physics that governs its response;
(ii)~\textbf{HOI Agent}, which writes how the hands touch these objects and move them by contact under the physics of each kind of matter;
and (iii)~\textbf{Evaluator Agent}, which verifies that each result obeys physical laws.
The HOI Agent specifies each hand by where it touches an object, and each object is simulated with the established physics solver of its matter~\cite{hu2018mlsmpm,sifakis2012fem,zhu2005sand,macklin2016xpbd}.
When a scene mixes several kinds of matter, one solver passes its state to another so that they can interact.

These agents serve reconstruction and generation alike, recovering the interaction shown in a video or producing one that follows an instruction from an image, a 3D scene or text.
For evaluation, we introduce \bench with 20 real-world reconstruction clips and 80 generation tasks across rigid, deformable, co-dimensional and flowing matter~\cite{shen2026fourdcodebench}.
Its tasks span elastic, plastic, granular and fluid behavior and include multi-physics cases, so success depends on how each material responds.
Since every object keeps its state and material in physical units, we further transfer the results to an independent physics engine~\cite{genesis2024}.
There, dexterous robot hands~\cite{sharpa2025wave,haddadin2022franka} track our hand motion without a learned policy while the engine simulates each object.
This transfer shows that our interactions remain plausible beyond our own kinematic hands and solvers.

Our contributions are as follows:
\begin{itemize}[leftmargin=*,itemsep=1pt,topsep=2pt]
  \item We present \method, to our knowledge the first agentic framework for physically plausible 4D HOI. Unlike prior coding agents, it writes the hands themselves and lets their physical contact drive the interaction.
  \item We introduce \bench with 100 reconstruction and generation tasks spanning four kinds of matter, on which \method writes plausible interactions for every kind with off-the-shelf frontier models.
  \item Every object carries its own physics on an established solver, so our interactions hold beyond our own simulation. We show qualitatively that they transfer to dexterous robot hands in an independent physics engine as potential demonstrations for robot learning.
\end{itemize}

%% file: sec/2_Related_work.tex
\section{Related Work}
\label{sec:related}

\paragraph{Hand-object reconstruction.}
Reconstructing hands and the objects they manipulate from images or videos has been widely studied.
Learning-based methods train on captured hand-object datasets~\cite{hampali2020honnotate,chao2021dexycb,yang2022oakink,fan2023arctic,zhan2024oakink2,kwon2021h2o,wang2024hocap,fu2025gigahands} to regress a parametric MANO~\cite{romero2017mano} hand from a single image~\cite{pavlakos2024hamer,potamias2025wilor} or jointly with the object it holds~\cite{hasson2019obman,ye2022ihoi,chen2025hort}.
Optimization-based methods fit hands and objects to observations under contact and shape priors~\cite{cao2021reconstructing,ye2023diffhoi,fan2024hold}, and recent work draws on large foundation models~\cite{liu2025easyhoi,dwivedi2025interactvlm} or recovers 4D interactions from video~\cite{xie2026cari4d}.
These methods assume rigid or articulated objects. Those with a simulator in the loop extend to deformable ones~\cite{jiang2025phystwin,lee2026physhandi}, yet reconstruct a video with a single material model driven by tracked hands.
\method instead reconstructs interactions as programs in which hands move each object by contact under the physics of its matter, so a single scene can combine several kinds of matter, from rigid bodies to liquids.

\paragraph{Hand-object interaction generation.}
Methods that synthesize new interactions rather than recover observed ones have progressed from static grasps to full motions.
Grasp generators predict hands that hold a given object~\cite{taheri2020grab,jiang2021graspTTA}, increasingly conditioned on language through semantic grasps or contact maps~\cite{li2024semgrasp,zhang2024nl2contact}.
Diffusion models generate hand-object motion from text~\cite{cha2024text2hoi,christen2024diffh2o,prakash2025how}, and physics-based methods learn policies that grasp and move objects in a simulator~\cite{christen2022dgrasp}.
Recent work further grounds affordances with multimodal language models~\cite{zhang2025openhoi}, lifts a photograph and an instruction to a 4D interaction~\cite{zhang2026photohoi} or enforces constraints such as avoiding a part during sampling~\cite{kim2026potential}.
Trained on captures of rigid or articulated objects, these models rarely extend to objects that deform, drape or flow.
\method instead requires no training data, as its agents write interactions with such objects from an image, a 3D scene or text and verify them against physical laws.

\paragraph{Coding agents for 3D and 4D content.}
Coding agents write, run and revise programs that create visual content.
They build scenes~\cite{hu2024scenecraft,yang2024holodeck,huang2024blenderalchemy,lu2025ll3m}, shapes~\cite{dai2025meshcoder,he2026thinking} and articulated objects~\cite{mandi2024real2code,le2025articulate} as code that preserves real size, parts and joints, unlike image-to-3D generators~\cite{xiang2025trellis,hunyuan2025hunyuan3d,chen2025sam3d,lin2025partcrafter}.
Agents also reconstruct objects from video~\cite{mazur2026agentstar} and reproduce dynamic scenes in executable benchmarks~\cite{shen2026fourdcodebench,tang2026video2world,shao2026physcodegym}, often with physics simulation.
They refine their programs through feedback on the results~\cite{gu2025blendergym,shinn2023reflexion,madaan2023selfrefine,huang2024cannot,wang2024voyager}, which frequently comes from model judges that favor certain answer positions and their own outputs~\cite{zheng2023judging,panickssery2024llm}.
Yet none of them models a hand, whose physical contact with an object drives the interaction while the material dictates the response.
\method brings such agents to hand-object interaction, where a program places each hand by its contacts and simulates each object under the physics of its matter.
It also grounds its feedback in physical measurements of each result and in bias-guarded model judges.

\paragraph{Physics simulation.}
Physically based animation models how each kind of matter responds to forces with established solvers, including MPM~\cite{hu2018mlsmpm} for elastoplastic and granular materials, FEM~\cite{sifakis2012fem} for elastic solids, FLIP~\cite{zhu2005sand} for liquids and XPBD~\cite{macklin2016xpbd} for cloth and rope.
General-purpose engines integrate several solvers~\cite{genesis2024}, and vision methods fit them to deformable objects in video~\cite{jiang2025phystwin,yin2026physcore,lee2026physhandi}.
Agents use simulators to plan robot actions~\cite{liu2026simpact} or generate robot learning tasks~\cite{wang2023robogen}, and some write simulations to reproduce a video~\cite{shen2026fourdcodebench} but omit the hand.
Simulated hands learn dynamic grasps~\cite{christen2022dgrasp} or imitate human videos~\cite{qin2022dexmv}, yet only on rigid bodies.
\method builds on the established solvers and combines them in one program, where each object follows the solver of its matter and the hands act as moving colliders.
Its results also transfer to dexterous robot hands~\cite{sharpa2025wave,haddadin2022franka} in an independent engine~\cite{genesis2024} across all four kinds of matter.

%% file: sec/3_Method.tex
\section{\method}
\label{sec:method}

\begin{figure*}[t]
  \centering
  \includegraphics[width=\linewidth]{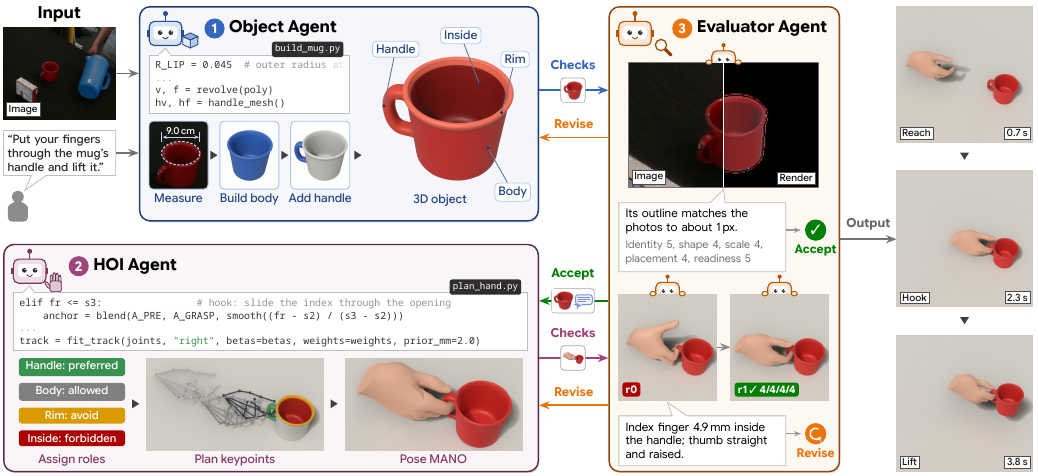}
  \caption{Overview of \method.
  From an input and an instruction, three agents write a 4D hand-object interaction as a program.
  (i)~The \textcolor{figObj}{Object Agent} builds every object in the input and gives it named parts, joints and its own physics.
  (ii)~The \textcolor{figHOI}{HOI Agent} decides which parts the hands should touch or avoid, defines each hand through its contacts and simulates how every object responds.
  (iii)~The \textcolor{figEval}{Evaluator Agent} checks each result by measurement and by its renders, then accepts it or sends it back with a review.
  Here round~0 is returned because the index finger penetrates the handle and the thumb is raised, and the revised round~1 is accepted.}
  \label{fig:overview}
\end{figure*}

\method represents each interaction as a program that can be run and verified.
Given an input $\mathcal{X}$ and an instruction $\ell$, the task is to write a self-contained program $P$ that produces
\begin{equation}
  P(\mathcal{X},\ell) \;\rightarrow\; \bigl(\{\mathcal O_t,\ \mathcal M_t\}_{t=1}^{T},\ \Pi,\ I_{1:T}\bigr),
  \label{eq:task}
\end{equation}
namely object meshes $\mathcal O_t$ and hands $\mathcal M_t$ over $T$ frames, a camera $\Pi$ and a render $I_{1:T}$.
Each hand is a parametric MANO~\cite{romero2017mano} model defined by its pose and translation.
In reconstruction, $\mathcal X$ is a video to match frame by frame, whereas in generation it is an image, a 3D scene or a text description.
The program determines how the hands interact with objects and how each object responds under its own physics, while the hand model is fixed so that no agent can shrink a hand to avoid collisions.
The result must be physically plausible and faithful to the input or instruction.

\method solves this task in three stages (see \Cref{fig:overview}), each run by an agent with its own context, instructions and tools.
(i)~The \textbf{Object Agent} (\Cref{sec:method:object}) builds the scene at rest, reconstructing or creating every object at real size with named parts, joints and the physics of its matter.
(ii)~The \textbf{HOI Agent} (\Cref{sec:method:hoi}) then writes the physical interaction, placing the hands by the contacts it plans on these parts and letting every object respond under its own physics.
(iii)~The \textbf{Evaluator Agent} (\Cref{sec:method:eval}) verifies both stages against physical laws and the input or instruction by measuring each result and inspecting its renders.
Each stage is a closed loop in which a generator $\mathcal G$ revises its program $P_r$ of round $r$ after an evaluator $\mathcal E$ assesses it,
\begin{equation}
  P_{r+1}=\mathcal G\bigl(P_r,\,\mathcal E(P_r)\bigr).
  \label{eq:loop}
\end{equation}
The loop ends when the evaluator accepts a round or the revision budget is exhausted, and it returns the best round.

\subsection{Object Agent}
\label{sec:method:object}
In the first stage, the Object Agent builds the $N$ objects of the input as a scene at rest for the hands to act on.
It writes one procedural modeling program per object, and together these programs produce
\begin{equation}
  P^{o}(\mathcal X,\ell)\;\rightarrow\;\bigl(\{\mathcal O^n\}_{n=1}^{N},\ \Pi\bigr),
  \label{eq:object}
\end{equation}
where each object $\mathcal O^n=(\mathcal S^n,\mathcal J^n,\kappa^n,\boldsymbol\xi^n)$ consists of a closed metric mesh $\mathcal S^n$ split into named parts, joints $\mathcal J^n$, a kind of matter $\kappa^n$ and material parameters $\boldsymbol\xi^n$, and $\Pi$ is the camera.
Named parts specify where a hand may or may not touch, and joints with axes and limits let articulated parts move as the real ones do.
The kind of matter $\kappa^n$ (\ie, rigid, deformable, co-dimensional or flowing) selects how the object is simulated, and $\boldsymbol\xi^n$ sets its parameters in physical units initialized from material presets.
Physics thus becomes a property of the object itself that governs its interaction with the hands in the next stage.
An object critic serves as $\mathcal E$ in Eqn.~\eqref{eq:loop} and compares renders from all sides and at the joint limits against the input for identity, shape, scale, placement and readiness for interaction.
The agent revises $P^{o}$ until the critic judges the scene faithful to the input.

\subsection{HOI Agent}
\label{sec:method:hoi}
The core of \method is the HOI Agent, which writes the hands that prior coding agents omit.
In its program $P$ of Eqn.~\eqref{eq:task}, the hands $\mathcal M_t$ \emph{interact} with objects $\mathcal O_t$ only through contact, and each object responds under its own physics.
The agent first plans which named parts a hand should touch or keep clear of and splits the motion into steps with goals.
Writing $P$ then requires two things that a scene at rest does not.
The \emph{physics} must let each kind of matter respond as its material dictates, and the \emph{hands} must coordinate many joints to make the planned contacts at the right time.

A rigid or articulated object moves with the hand that grasps it or under rigid-body dynamics~\cite{coumans2015bullet}.
Non-rigid matter responds in ways that keyframes rarely reproduce, so the agent simulates it as its $\kappa^n$ and $\boldsymbol\xi^n$ specify.
Like general-purpose engines~\cite{genesis2024}, we run the established solver of each regime in Warp~\cite{macklin2022warp} with the hands as frictional kinematic colliders:
(i)~\textbf{MPM}~\cite{hu2018mlsmpm} evolves particle deformation gradients $\mathbf F_p\leftarrow(\mathbf I+\Delta t\,\nabla\mathbf v_p)\mathbf F_p$ from velocities $\mathbf v_p$ at time step $\Delta t$ on a background grid under elastic, plastic, granular or viscous laws for matter that loses its rest shape;
(ii)~\textbf{FEM}~\cite{sifakis2012fem} evaluates the elastic energy $\psi(\mathbf F)$ on the object's own tetrahedra, so elastic solids recover their rest shape with an intact surface;
(iii)~\textbf{FLIP}~\cite{zhu2005sand} projects the velocity $\mathbf v$ to $\nabla\!\cdot\mathbf v=0$, sparing thin liquids the small time steps of a compressible fluid;
and (iv)~\textbf{XPBD}~\cite{macklin2016xpbd} projects stretch constraints $C_j(\mathbf x)=0$ on positions $\mathbf x$ with compliance $\alpha_j$ and adds bending~\cite{bergou2006bending} for cloth and rope, thinner than a grid cell.
The four solvers run together in one scene with the same hands, and one can act on another as a collider, source or sink.
A single program thus expresses multi-physics interactions, such as a towel soaking up spilled water.
The agent may tune parameters, combine solvers and add colliders, but every deformation must come from a solver.
A result is accepted only if a replay reproduces it, so all motion comes from simulation.
The simulation is plausible rather than predictive~\cite{barzel1996plausible}, as materials start from presets and forces pass one way, from hands to objects and between solvers.

The hands must establish the contacts that drive the object's response.
Since contact governs the interaction, the agent specifies each hand by its contacts rather than its pose.
It prescribes where the fingers and palm should lie on and around the object over time~\cite{huang2024rekep} and fits the MANO hands $\mathcal M_t$ to the prescribed positions with natural, smooth and anatomically valid motion.
In reconstruction, the hands detected in the video~\cite{jiang2023rtmpose,khirodkar2026sapiens2} initialize the fitting.
A subsequent grasp refinement closes the remaining gaps so that the fingers rest on the surface without penetration~\cite{barill2018fastwinding}.
The agent thus decides where and when a hand makes contact, while the fitting determines the articulation that achieves it.
\Cref{app:details} details the planner, the solvers and the hand fitting.

\begin{figure*}[t]
  \centering
  \includegraphics[width=\linewidth]{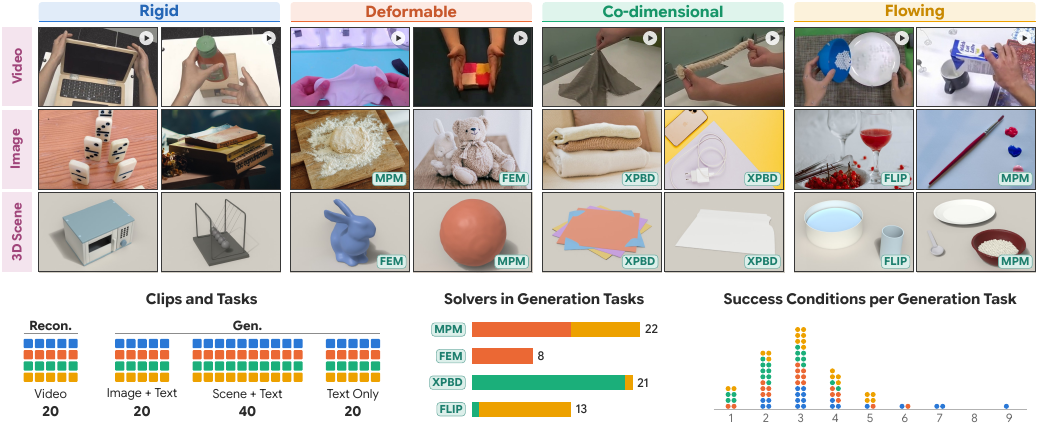}
  \caption{Overview of \bench, with 20 reconstruction clips and 80 generation tasks balanced across four kinds of matter.
  (Top) Two examples per kind of matter for each visual input, \ie, real-world videos for reconstruction and images and 3D scenes for generation, with tags naming the solver each non-rigid example requires.
  (Left) Clips and tasks per input, one square each.
  (Middle) Generation tasks per solver, where the four multi-physics tasks count toward both of their solvers.
  (Right) Distribution of generation tasks over the number of hidden success conditions, one marker per task.
  Squares, bars and markers are colored by kind of matter as in the column headers.}
  \label{fig:bench}
\end{figure*}
\subsection{Evaluator Agent}
\label{sec:method:eval}
The Evaluator Agent serves as $\mathcal E$ in Eqn.~\eqref{eq:loop} for both stages and determines whether each result is physically plausible and faithful to the input or instruction.
For the interaction $P$ of Eqn.~\eqref{eq:task}, it returns the feedback $\mathcal E(P_r)=(\mathcal V_r,\mathbf s_r,c_r)$ from a \emph{checker} and a \emph{critic}.
The violations $\mathcal V_r$ list where the result breaks a physical law or misses a planned goal, as measured by the deterministic checker.
From renders, the critic scores the hands, contacts, objects and motion in $\mathbf s_r$ and states what to fix in the review $c_r$.
Physical laws are thus checked by measurement rather than by a model's impression.
Together, the three components tell the generator whether a round fails and how to revise it.

The checker measures the result $\{\mathcal O_t,\mathcal M_t\}_{t=1}^{T}$ before the critic sees it.
It verifies physical plausibility as necessary conditions on contact, carrying and support, namely that no hand penetrates matter, that forces within friction cones~\cite{ferrari1992planning} can hold what is carried and that nothing moves untouched or floats.
It also checks hand validity, part roles, planned goals and, in reconstruction, agreement with the video.
The critic then inspects renders alongside the measurements and assesses what they miss (\eg, a thumb twisted to its limit or paper that drapes like cloth).
It receives only the submitted result in a fresh context, which prevents anchoring on earlier attempts and lenient grading~\cite{laban2026lost,pan2024spontaneous}.
Checker warnings cap the related scores, so the critic cannot accept what the measurements contradict.
The critic may also revise a faulty plan.
A round is accepted only when $\mathcal V_r$ is empty, every score in $\mathbf s_r$ is acceptable and $c_r$ raises no major issue.

%% file: sec/4_Evaluation.tex
\section{Evaluation}
\label{sec:exp}

\begin{table*}[t]
  \centering
  \caption{4D HOI reconstruction on \bench with each model as the backbone of \method.
  Each kind of matter has five clips evaluated per frame against the GT of their source datasets, and the last two columns cover all 20 clips.
  Each metric is averaged over the clips that annotate it.
  The best results are shown in \textbf{bold} and the second best are \underline{underlined}.}
  \label{tab:video}
  \vspace{-0.7em}
  \setlength{\tabcolsep}{3pt}
  \resizebox{\linewidth}{!}{%
  \begin{tabular}{l|cccc|cccc|cccc|cccc|cc}
    \toprule
    & \multicolumn{4}{c|}{\textcolor{matRigid}{\textbf{Rigid}}} & \multicolumn{4}{c|}{\textcolor{matDeform}{\textbf{Deformable}}} & \multicolumn{4}{c|}{\textcolor{matCodim}{\textbf{Co-dimensional}}} & \multicolumn{4}{c|}{\textcolor{matFlow}{\textbf{Flowing}}} & \multicolumn{2}{c}{\textbf{All}} \\
    \cmidrule(lr){2-5}\cmidrule(lr){6-9}\cmidrule(lr){10-13}\cmidrule(lr){14-17}\cmidrule(lr){18-19}
    \multicolumn{1}{c|}{\textbf{Model}} & MJE ($\downarrow$) & CD$_o$ ($\downarrow$) & MRRPE ($\downarrow$) & C-F1 ($\uparrow$) & MJE ($\downarrow$) & CD ($\downarrow$) & IoU$_\text{v}$ ($\uparrow$) & Vol. ($\downarrow$) & MJE ($\downarrow$) & CD ($\downarrow$) & Track ($\downarrow$) & IoU$_\text{h}$ ($\uparrow$) & MJE ($\downarrow$) & CD$_o$ ($\downarrow$) & C-F1 ($\uparrow$) & In ($\uparrow$) & Pen. ($\downarrow$) & Acc. ($\downarrow$) \\
    \midrule
    \modelicon{claude}Claude Opus 5.5 & 29.1 & \underline{59.4} & 52.0 & 0.35 & 26.0 & 43.0 & 0.60 & 10.9 & 27.9 & 44.6 & 74.3 & 0.34 & 30.3 & 74.2 & 0.37 & \textbf{21.4} & 1.0 & \textbf{1.6} \\
    \modelicon{claude}Claude Sonnet 5.5 & 27.1 & 81.4 & \underline{48.4} & 0.31 & 31.6 & 55.1 & 0.55 & \phantom{0}\textbf{3.1} & \underline{27.5} & 55.7 & 69.9 & 0.33 & 28.3 & \underline{52.2} & 0.30 & \phantom{0}9.0 & 1.0 & \underline{1.7} \\
    \modelicon{chatgpt}GPT-6 Astra & \textbf{26.0} & 62.9 & 49.0 & \underline{0.44} & \underline{24.3} & \underline{38.7} & \textbf{0.63} & \phantom{0}\underline{6.3} & 33.4 & \underline{39.6} & \underline{45.8} & \underline{0.43} & \textbf{21.7} & 58.3 & \underline{0.45} & 15.7 & \underline{0.7} & \textbf{1.6} \\
    \modelicon{chatgpt}GPT-6.1 Sol & \underline{26.4} & \textbf{43.9} & \textbf{47.7} & \textbf{0.45} & \textbf{21.7} & \textbf{21.0} & \underline{0.61} & 12.5 & \textbf{17.6} & \textbf{36.8} & \textbf{44.6} & \textbf{0.44} & \underline{24.2} & \textbf{51.3} & \textbf{0.48} & \underline{18.5} & \textbf{0.6} & \underline{1.7} \\
    \bottomrule
  \end{tabular}}
\end{table*}
\begin{table*}[t]
  \centering
  \caption{4D HOI generation on \bench with each model as the backbone of \method.
  Each kind of matter pools its 20 tasks over image, 3D scene and text inputs.
  The best results are shown in \textbf{bold} and the second best are \underline{underlined}.}
  \label{tab:gen}
  \vspace{-0.7em}
  \setlength{\tabcolsep}{4pt}
  \resizebox{\linewidth}{!}{%
  \begin{tabular}{l|cccc|cccc|cccc|cccc}
    \toprule
    & \multicolumn{4}{c|}{\textcolor{matRigid}{\textbf{Rigid}}} & \multicolumn{4}{c|}{\textcolor{matDeform}{\textbf{Deformable}}} & \multicolumn{4}{c|}{\textcolor{matCodim}{\textbf{Co-dimensional}}} & \multicolumn{4}{c}{\textcolor{matFlow}{\textbf{Flowing}}} \\
    \cmidrule(lr){2-5}\cmidrule(lr){6-9}\cmidrule(lr){10-13}\cmidrule(lr){14-17}
    \multicolumn{1}{c|}{\textbf{Model}} & SR ($\uparrow$) & S ($\uparrow$) & Phys ($\uparrow$) & Elo ($\uparrow$) & SR ($\uparrow$) & S ($\uparrow$) & Phys ($\uparrow$) & Elo ($\uparrow$) & SR ($\uparrow$) & S ($\uparrow$) & Phys ($\uparrow$) & Elo ($\uparrow$) & SR ($\uparrow$) & S ($\uparrow$) & Phys ($\uparrow$) & Elo ($\uparrow$) \\
    \midrule
    \modelicon{claude}Claude Opus 5.5 & 70.0 & 0.89 & \textbf{0.91} & \phantom{0}963.9 & \underline{55.0} & \underline{0.85} & 0.80 & \phantom{0}951.0 & \textbf{60.0} & \textbf{0.80} & \textbf{0.93} & \phantom{0}982.8 & \underline{60.0} & \textbf{0.91} & 0.78 & \phantom{0}920.9 \\
    \modelicon{claude}Claude Sonnet 5.5 & 65.0 & 0.94 & 0.76 & \phantom{0}888.5 & 40.0 & 0.78 & 0.73 & \phantom{0}923.6 & 50.0 & \underline{0.79} & 0.73 & \phantom{0}874.0 & 50.0 & 0.85 & 0.82 & \phantom{0}896.6 \\
    \modelicon{chatgpt}GPT-6 Astra & \textbf{90.0} & \underline{0.98} & \underline{0.90} & \textbf{1102.0} & \textbf{70.0} & \textbf{0.91} & \textbf{0.96} & \underline{1058.1} & \underline{55.0} & 0.78 & 0.85 & \underline{1069.3} & \textbf{80.0} & \underline{0.87} & \underline{0.99} & \textbf{1098.5} \\
    \modelicon{chatgpt}GPT-6.1 Sol & \underline{85.0} & \textbf{1.00} & 0.84 & \underline{1045.6} & \underline{55.0} & \underline{0.85} & \underline{0.89} & \textbf{1067.3} & \underline{55.0} & \textbf{0.80} & \underline{0.88} & \textbf{1074.0} & \underline{60.0} & 0.79 & \textbf{1.00} & \underline{1084.0} \\
    \bottomrule
  \end{tabular}}
\end{table*}
\subsection{\texorpdfstring{\bench}{CodeHOI-Bench}}
\label{sec:exp:bench}
\paragraph{Tasks.}
\bench comprises two tasks over rigid, deformable, co-dimensional and flowing matter~\cite{shen2026fourdcodebench}.
(i)~\textbf{4D HOI Reconstruction} recovers the interaction shown in a video as hands, part-aware object meshes and the state of the matter in every frame.
Its 20 clips are drawn five per kind of matter from seven real-world HOI datasets~\cite{fan2023arctic,chao2021dexycb,yin2026physcore,fu2025gigahands,jiang2025phystwin,zhan2024oakink2,kwon2021h2o} and span articulated objects, dough, plush toys, cloth, rope, grains and liquid.
(ii)~\textbf{4D HOI Generation} creates an interaction that follows an instruction on objects given as an image, a 3D scene or text alone.
Its 80 tasks with 20 per kind of matter hinge on how the matter responds, and pairs such as water or honey in one cup call for different motions on identical geometry.
\Cref{fig:bench} summarizes the inputs, kinds of matter, solvers and success conditions.

\paragraph{Evaluation protocols.}
All metrics use the submitted meshes and hands, while only the scorer sees the ground truth (GT) and hidden labels.
\textbf{Reconstruction} is evaluated against the GT per frame.
Hands are scored by mean joint error (MJE) for pose and acceleration error (Acc.) for jitter~\cite{fan2023arctic}.
Rigid objects are scored by object Chamfer distance (CD$_o$), the relative hand-object position error (MRRPE) and contact F1 (C-F1)~\cite{xie2026cari4d,fan2023arctic}.
Non-rigid matter is scored by Chamfer distance (CD) to observed points, the drift of tracked points (Track), silhouette IoU in the input (IoU$_\text{v}$) or held-out views (IoU$_\text{h}$) and the peak volume change (Vol.)~\cite{jiang2025phystwin,shen2026fourdcodebench}.
We add the containment of poured matter (In) and hand-object penetration depth (Pen.).
\textbf{Generation} lacks reference motion, so each task hides success conditions and part roles from agents (\eg, the lid must end off the pot and be grasped by its knob).
A task counts as solved (SR) only if all conditions hold without physical violation~\cite{wang2023dexgraspnet}.
We also report the fraction of conditions met ($S$)~\cite{srivastava2021behavior}, the worst physical check score (Phys) and the Elo rating~\cite{elo1978rating,chiang2024chatbot} from a judge's pairwise visual comparisons.
The judge is blind to the producing agent~\cite{panickssery2024llm} and sees both orders of each pair~\cite{zheng2023judging}.
\Cref{app:details} details the metrics and the judge.

\subsection{Experimental Details}
\label{sec:exp:setup}
\paragraph{Implementation details.}
Each model runs in its provider's agent framework~\cite{anthropic2025claudecode,openai2025codex} in an isolated environment with one GPU and no web access.
All three agents of a run share one model and reasoning effort.
Each stage of Eqn.~\eqref{eq:loop} allows at most three revisions and stops early after two rounds without improvement.
The Object Agent has 40 minutes for its first round and 25 minutes per revision in both tasks, and the HOI Agent receives 10 additional minutes each as it also runs the simulations.
The object critic scores identity, shape, scale, placement and readiness for interaction, and the interaction critic scores hand naturalness, contact, objects and motion.
Each criterion ranges from 1 to 5, with 3 or higher deemed acceptable.
\Cref{app:details} details the environment and the checker, and the \href{https://shpark97.github.io/CodeHOI}{code} will be released upon publication.

\paragraph{Baselines.}
No prior method both reconstructs and generates physically plausible HOI across all four kinds of matter that \bench covers.
We therefore compare four frontier models as the backbone of \method, namely Opus~5.5, Sonnet~5.5, GPT-6 Astra and GPT-6.1 Sol~\cite{anthropic2025claudecode,openai2025codex}.
All four receive the same instructions and tools, so the results differ only by the model.
Each backbone runs once per clip and task at \emph{high} reasoning effort.

\subsection{Main Results}
\label{sec:exp:main}
\begin{figure*}[t]
  \centering
  \includegraphics[width=\linewidth]{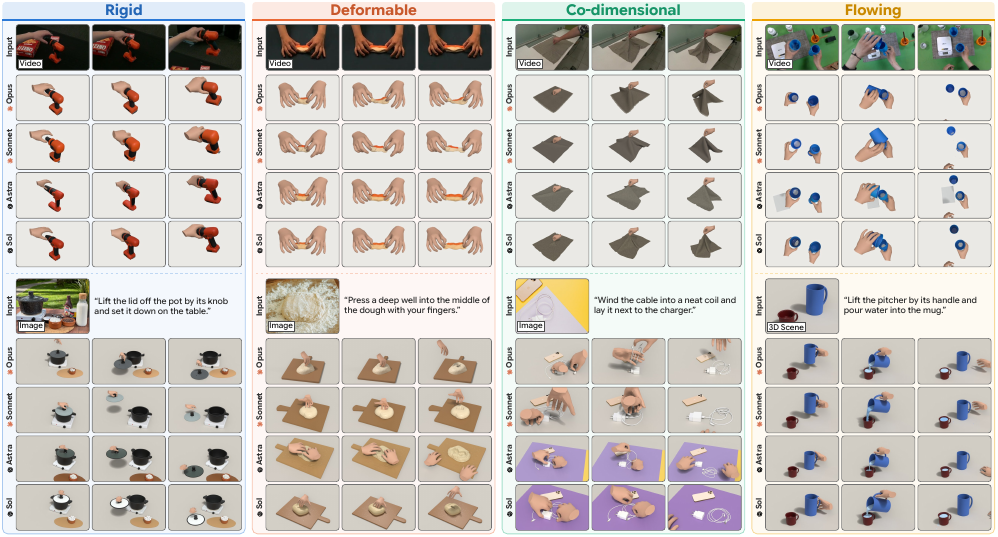}
  \caption{Qualitative comparison of the four backbones on \bench.
  Each column covers one kind of matter, with a reconstruction clip in the upper block and a generation task in the lower block.
  In each block, the input row shows three frames of the video or the image or 3D scene with its instruction, and the rows below show three frames of the interaction written with each backbone (\ie, Opus~5.5, Sonnet~5.5, GPT-6 Astra and GPT-6.1 Sol).
  Reconstructions follow the input closely for every backbone, whereas generations differ more in the objects each backbone builds and in how it carries out the instruction.}
  \label{fig:qual}
\end{figure*}
\paragraph{4D HOI reconstruction.}
\label{sec:exp:video}
\Cref{tab:video} reports reconstruction on the 20 clips for each backbone.
\mbox{GPT-6.1} Sol leads in 12 of the 18 metrics, yet all four backbones share similar strengths and weaknesses.
On rigid clips, the hands differ little across backbones with an MJE of 26.0 to 29.1, since every fit starts from the same detected hands. Object accuracy varies more, as CD$_o$ ranges from 43.9 to 81.4 depending on how closely each program matches the real shape and scale.
Contact is the weakest aspect, as \mbox{C-F1} stays at or below 0.48 even though penetration remains within 1.0\,mm.
The hands thus avoid the object but miss where the GT hands touch it.
Deformable and co-dimensional matter separates the backbones most, as deformable CD ranges from 21.0 to 55.1 and the GPT backbones track co-dimensional matter with a Track near 45 against about 70 for the Claude backbones.
Co-dimensional MJE spreads from 17.6 to 33.4 but only from 9.9 to 14.1 after Procrustes alignment (see \Cref{app:analyses}), so most hand error lies in global rotation and scale rather than articulation.
Flowing matter is the hardest, as containment stays at or below 21.4\% although the containers are recovered as well as rigid ones.
\method thus recovers trackable surfaces well, while exact contact and flowing matter remain open.

\paragraph{4D HOI generation.}
Here success is judged by hidden conditions, as \Cref{tab:gen} reports for the 80 tasks.
Every backbone solves rigid tasks most often, with an SR of 65.0 to 90.0 against 40.0 to 80.0 on non-rigid matter.
Failures differ in nature.
More than half of the unsolved rigid results meet every condition but fail a part role or a physical check, so the difficulty lies in how the hand acts.
About four in five unsolved non-rigid results instead leave a condition unmet, so the difficulty lies in reaching the required state of the matter.
Physical plausibility also depends on the backbone, since all share the same solvers yet Phys on deformable and flowing matter ranges from 0.73 to 0.82 for the Claude backbones and from 0.89 to 1.00 for the GPT backbones.
\mbox{GPT-6} Astra attains the highest SR on three kinds of matter, Opus~5.5 leads co-dimensional tasks in SR and Phys and Sonnet~5.5 trails on all four.
Elo still ranks both GPT backbones above both Claude backbones in every kind of matter, so task success and visual quality are distinct.
\Cref{fig:qual} compares the four backbones on one clip and one task per kind of matter.

\subsection{Transfer to Simulation}
\label{sec:exp:sim}
A physically plausible interaction should not depend on the particular hands and solvers that produced it.
Since each object carries its own physics in physical units, we test this by transferring our results to the independent physics engine Genesis~\cite{genesis2024}.
The test is demanding, since an actuated hand must replace our kinematic one at the same contacts and a different engine must reproduce the matter's response.
The MANO~\cite{romero2017mano} motion is retargeted~\cite{qin2022dexmv} to a \mbox{22-DoF} SharpaWave hand~\cite{sharpa2025wave} on a \mbox{7-DoF} Franka arm~\cite{haddadin2022franka} so that its fingertips keep the written contacts.
Without a learned policy, the robot tracks this motion with position control.
Position control alone cannot regulate grip force, so we simply attach an object or cloth region touched during a reference grasp to the hand until the reference releases it.

\begin{figure*}[t]
  \centering
  \includegraphics[width=\linewidth]{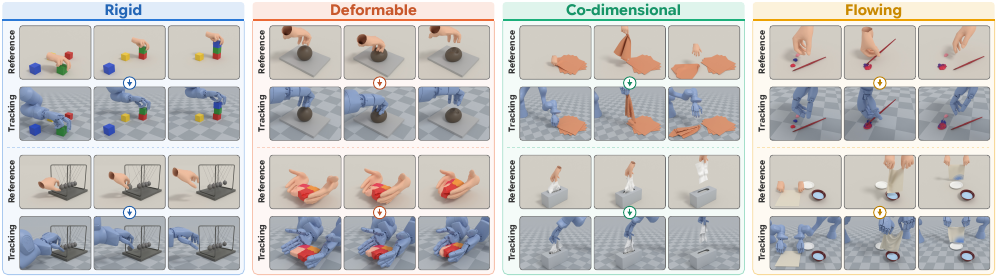}
  \caption{Transfer of \method results to dexterous SharpaWave hands~\cite{sharpa2025wave} on Franka arms~\cite{haddadin2022franka} in Genesis~\cite{genesis2024}.
  For two results per kind of matter, each reference from \method is shown above its tracking in simulation at three frames.
  Without a learned policy, the robot hands follow the retargeted motion and reach the written contacts on time, while Genesis simulates each object with the solver of its matter.
  A few minor adjustments thus suffice for the interactions to carry over to all four kinds of matter.}
  \label{fig:sim}
\end{figure*}

\Cref{fig:sim} shows two transferred results per kind of matter.
Although larger than human hands, the robot hands establish the written contacts on time.
As in the reference, a released ball of a Newton's cradle drives the last one outward, pressed dough deforms by contact alone and a cloth lifted from a stack drapes as it is set down.
With only minor adjustments, our interactions thus transfer plausibly across diverse matter.
Transfer without them remains hard, as retargeting across embodiments is itself challenging~\cite{handa2020dexpilot,qin2022dexmv}.
These rollouts could in turn serve as demonstrations for reinforcement learning~\cite{rajeswaran2018dapg} or vision-language-action models~\cite{kim2024openvla}, which we leave to future work.
We show further transfers in \Cref{app:more_qual}.

\subsection{Further Analysis}
\label{sec:exp:analysis}

\begin{figure}[t]
  \centering
  \begin{subfigure}[t]{0.5\linewidth}
    \centering
    \includegraphics[width=\linewidth]{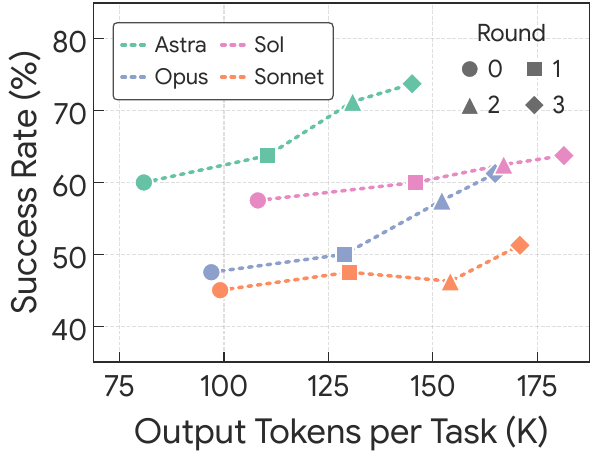}
    \caption{Token efficiency}
    \label{fig:analysis:cost:tokens}
  \end{subfigure}%
  \begin{subfigure}[t]{0.5\linewidth}
    \centering
    \includegraphics[width=\linewidth]{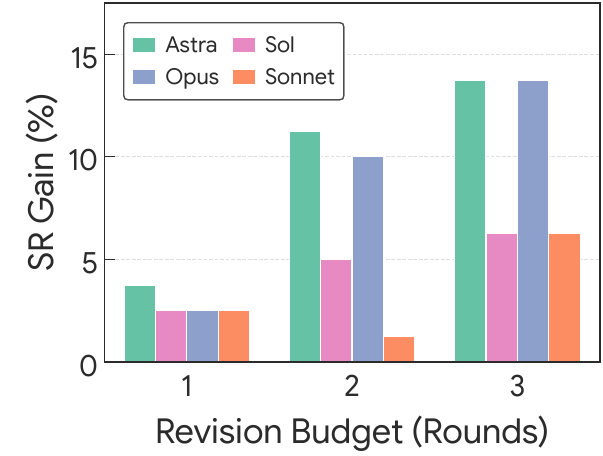}
    \caption{Revision gains}
    \label{fig:analysis:cost:budget}
  \end{subfigure}
  \par\vspace{-0.5em}
  \caption{Revision efficiency on the 80 generation tasks. (a) SR of the result retained by each backbone at each revision budget against the output tokens spent per task, with marker shapes denoting the round. (b) SR gain over the first round at each revision budget.}
  \label{fig:analysis:cost}
\end{figure}

\paragraph{Revision efficiency.}
\label{sec:exp:scaling}
The loop of Eqn.~\eqref{eq:loop} revises every result, so we isolate its effect by also scoring the first-round results, written with the checker before any critique.
These already solve 45.0 to 60.0\% of the 80 generation tasks, and \Cref{fig:analysis:cost} shows that revision raises SR by 6.3 to 13.8 points for every backbone at 1.7 to 1.8 times the output tokens.
Most of the final success thus stems from writing with the checker, and revision adds a modest gain at moderate cost.
The gain grows with the revision budget but unevenly across backbones.
One revision adds 2.5 to 3.8 points for every backbone, whereas three add 13.8 points for \mbox{GPT-6} Astra and Opus~5.5 but only 6.3 for Sonnet~5.5 and \mbox{GPT-6.1} Sol.
Revision also preserves the order of the backbones, as \mbox{GPT-6} Astra leads at every budget while spending the fewest tokens.
Its first round alone solves more tasks than three revisions of Sonnet~5.5 at less than half the tokens, so the choice of backbone matters more than the number of revisions.

\paragraph{Judge consistency.}
\label{sec:exp:judge}
The Elo ratings~\cite{elo1978rating} in \Cref{tab:gen} are assigned by Opus~5.5 as the sole judge, which may favor the results of its own model family~\cite{panickssery2024llm}.
To test whether the ranking depends on the judge, we hold the results fixed and let each of the four models in turn judge the same pairs on five generation tasks per kind of matter.
\Cref{tab:analysis:reliability} shows that every judge ranks both GPT backbones above both Claude backbones and Opus above Sonnet, while only Sonnet puts Sol above Astra.
Each judge tends to rate its own model family somewhat higher, yet this preference of up to about 50 Elo stays well below the gap of at least 86 Elo between the families.
Self-preference thus shifts the ratings without reordering the model families.
Single pairwise judgments are less stable, as they can flip with the presentation order and differ between judges on a notable share of pairs.
The ranking therefore emerges from many comparisons rather than from any single judgment, which makes Elo a robust summary of relative visual quality.
This supports reporting Elo from a single judge in \Cref{tab:gen}, as the order of the model families does not depend on that choice.

\begin{table}[t]
  \centering
  \caption{Judge consistency on 20 generation tasks, five per kind of matter. Each column lists the Elo ratings~\cite{elo1978rating} that one model, acting as the judge, assigns to the same fixed results of the four backbones. Every judge places both GPT backbones above both Claude backbones, so the family order holds regardless of the judge.}
  \label{tab:analysis:reliability}
  \vspace{-0.7em}
  \resizebox{\linewidth}{!}{%
  \begin{tabular}{l|cccc}
    \toprule
    \multicolumn{1}{c|}{} & \multicolumn{4}{c}{\textbf{Judge}} \\
    \cmidrule(lr){2-5}
    \multicolumn{1}{c|}{\textbf{Model}} & \modelicon{claude}Opus & \modelicon{claude}Sonnet & \modelicon{chatgpt}Astra & \modelicon{chatgpt}Sol \\
    \midrule
    \modelicon{claude}Claude Opus 5.5 & 963.9 & 954.7 & 934.4 & 932.1 \\
    \modelicon{claude}Claude Sonnet 5.5 & 883.3 & 883.1 & 875.3 & 838.0 \\
    \modelicon{chatgpt}GPT-6 Astra & \textbf{1102.4} & \underline{1078.7} & \textbf{1124.7} & \textbf{1156.6} \\
    \modelicon{chatgpt}GPT-6.1 Sol & \underline{1050.4} & \textbf{1083.5} & \underline{1065.6} & \underline{1073.3} \\
    \bottomrule
  \end{tabular}}
\end{table}

%% file: sec/5_Conclusion.tex
\section{Conclusion}
\label{sec:conclusion}
This paper addressed physically plausible 4D hand-object interaction, in which hands act only through contact and each object responds as its material dictates.
We introduced \method, an agentic framework whose three agents write such interactions as programs.
The Object Agent builds the objects in a scene and assigns each its own physics.
The HOI Agent specifies the hands by their contacts and simulates each object with the established physics solver of its matter.
The Evaluator Agent verifies each result against physical laws and accepts it or returns it with a review.
We further introduced \bench, which evaluates reconstruction and generation across four kinds of matter.
On this benchmark, \method writes plausible interactions for every kind with off-the-shelf frontier models.
Since every object carries its physics in physical units, these interactions transfer with minor adjustments to dexterous robot hands in an independent physics engine.
The transferred rollouts may serve as demonstrations for robot learning, which we leave to future work.
Writing the hand thus brings physically plausible HOI across diverse matter within reach of coding agents.

\FloatBarrier
\clearpage
\section*{LLM Usage Statement}
Apart from serving as the agents of \method, LLMs (Claude and Codex) assisted with coding, running experiments, creating figures, improving readability and formatting references.
The authors directed the research, reviewed all AI-assisted work and verified the code and results.
They take full responsibility for the paper, including its text, claims and artifacts.

%% file: sec/6_Supplementary.tex
\clearpage
\onecolumn
\setcounter{page}{1}
\begin{center}
\Large\textbf{\thetitle}\\[0.5em]
Supplementary Material
\end{center}
\vspace{0.5em}
\raggedbottom
\newsavebox{\suppwidthbox}
\newcommand{\suppfitwidth}[1]{\sbox{\suppwidthbox}{#1}%
  \ifdim\wd\suppwidthbox>\linewidth\resizebox{\linewidth}{!}{\usebox{\suppwidthbox}}%
  \else\usebox{\suppwidthbox}\fi}
\newcommand{\appfig}[1]{\par\vspace{0.6em}\noindent\begin{minipage}{\textwidth}\captionsetup{type=figure}\centering #1\end{minipage}\par}
\captionsetup[table]{font=small,skip=3pt}
\newcommand{\suppneedspace}[1]{\par\begingroup\dimen0=\pagegoal\advance\dimen0-\pagetotal\ifdim\dimen0<#1\relax\newpage\fi\endgroup}

\section{Additional Qualitative Results}
\label{app:more_qual}\label{app:qual}
This section provides additional qualitative results of \method.
We first compare the four backbones on \bench for reconstruction and generation of rigid, deformable, co-dimensional and flowing matter in \Cref{app:more_qual:bench}.
We then show in \Cref{app:more_qual:sim} how the resulting interactions transfer to robot hands.

\subsection{Results on \bench}
\label{app:more_qual:bench}
We run \method with four backbones (\ie, Opus~5.5, Sonnet~5.5, GPT-6 Astra and GPT-6.1 Sol) and compare their results.
All four share the same inputs, instructions and tools, so the differences between their results reflect the models.
For each kind of matter, \Cref{fig:app_video_rigid,fig:app_video_deformable,fig:app_video_codim,fig:app_video_flowing} show the reconstructions of four video clips and \Cref{fig:app_gen_rigid,fig:app_gen_deformable,fig:app_gen_codim,fig:app_gen_flowing} show the generations of four tasks.

\vfill
\appfig{
\includegraphics[width=0.91\textwidth,page=1]{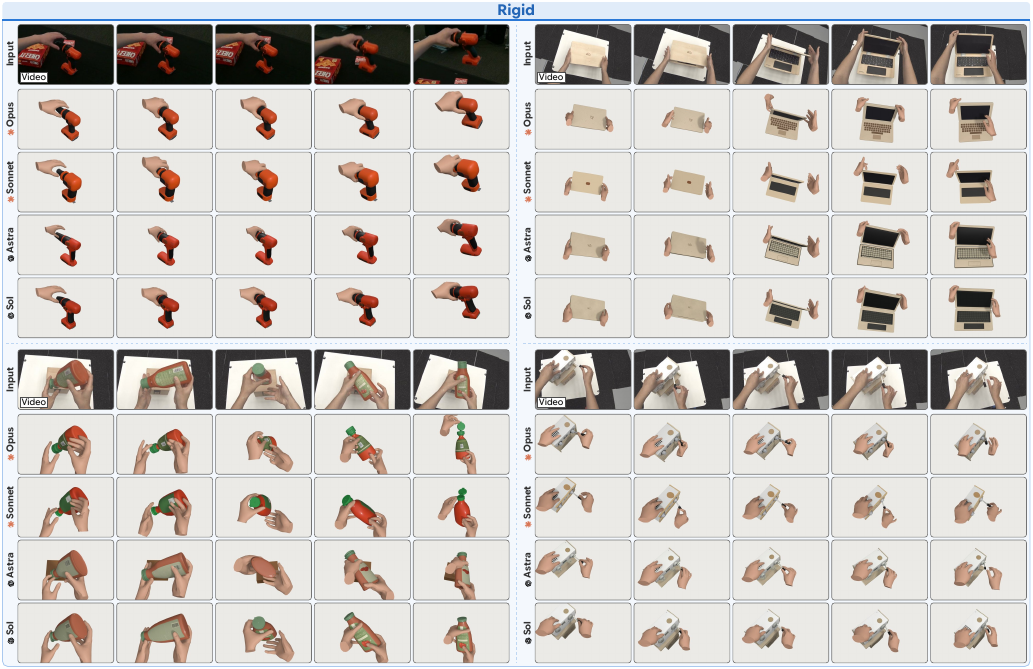}
\par\vspace{-0.5em}
\caption{Reconstruction of rigid clips of \bench by the four backbones.}
\label{fig:app_video_rigid}}
\vfill
\newpage

\vspace*{\fill}\vspace{-\dimexpr\topskip+0.6em+\parskip+8pt\relax}
\appfig{
\includegraphics[width=0.91\textwidth,page=2]{fig/supp_qual.pdf}
\par\vspace{-0.5em}
\caption{Reconstruction of deformable clips of \bench by the four backbones.}
\label{fig:app_video_deformable}}

\appfig{
\includegraphics[width=0.91\textwidth,page=3]{fig/supp_qual.pdf}
\par\vspace{-0.5em}
\caption{Reconstruction of co-dimensional clips of \bench by the four backbones.}
\label{fig:app_video_codim}}
\vspace*{\fill}
\newpage

\vspace*{\fill}\vspace{-\dimexpr\topskip+0.6em+\parskip+8pt\relax}
\appfig{
\includegraphics[width=0.91\textwidth,page=4]{fig/supp_qual.pdf}
\par\vspace{-0.5em}
\caption{Reconstruction of flowing clips of \bench by the four backbones.}
\label{fig:app_video_flowing}}

\appfig{
\includegraphics[width=0.91\textwidth,page=5]{fig/supp_qual.pdf}
\par\vspace{-0.5em}
\caption{Generation of rigid tasks of \bench by the four backbones.}
\label{fig:app_gen_rigid}}
\vspace*{\fill}
\newpage

\vspace*{\fill}\vspace{-\dimexpr\topskip+0.6em+\parskip+8pt\relax}
\appfig{
\includegraphics[width=0.91\textwidth,page=6]{fig/supp_qual.pdf}
\par\vspace{-0.5em}
\caption{Generation of deformable tasks of \bench by the four backbones.}
\label{fig:app_gen_deformable}}

\appfig{
\includegraphics[width=0.91\textwidth,page=7]{fig/supp_qual.pdf}
\par\vspace{-0.5em}
\caption{Generation of co-dimensional tasks of \bench by the four backbones.}
\label{fig:app_gen_codim}}
\vspace*{\fill}
\newpage

\appfig{
\includegraphics[width=0.91\textwidth,page=8]{fig/supp_qual.pdf}
\par\vspace{-0.5em}
\caption{Generation of flowing tasks of \bench by the four backbones.}
\label{fig:app_gen_flowing}}

\subsection{Transfer to Simulation}
\label{app:more_qual:sim}
We further transfer results of \method on \bench to dexterous robot hands~\cite{sharpa2025wave} on robot arms~\cite{haddadin2022franka} in Genesis~\cite{genesis2024} across diverse matter.
\Cref{fig:app_sim_1,fig:app_sim_2,fig:app_sim_3} show each \method reference above its tracking in simulation at the same frames.

\appfig{
\includegraphics[width=\linewidth,page=1]{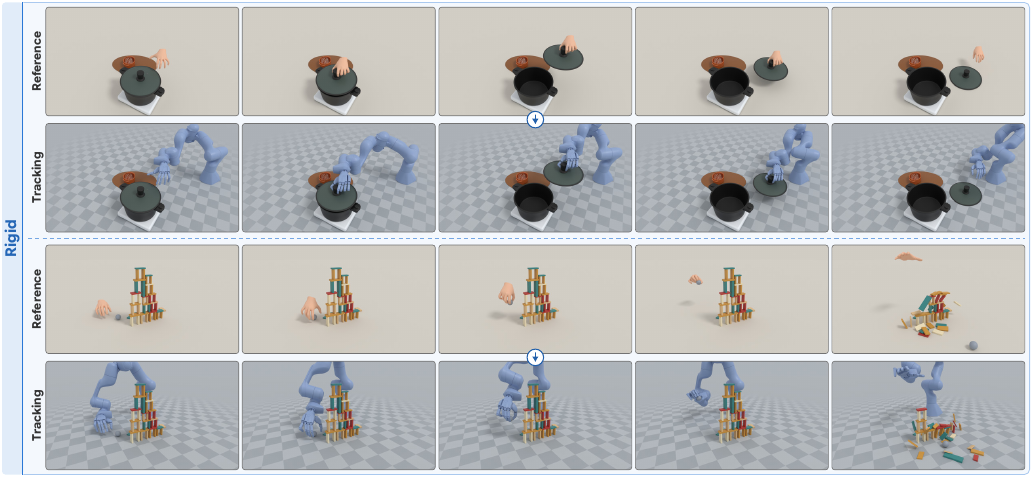}
\par\vspace{-0.5em}
\caption{Transfer of \method results on \bench to dexterous robot hands in Genesis.}
\label{fig:app_sim_1}}

\newpage
\vspace*{\fill}
\appfig{
\includegraphics[width=\linewidth,page=2]{fig/supp_sim.pdf}
\par\vspace{-0.5em}
\caption{Transfer of \method results on \bench to dexterous robot hands in Genesis (continued).}
\label{fig:app_sim_2}}
\vspace*{\fill}
\newpage

\vspace*{\fill}
\appfig{
\includegraphics[width=\linewidth,page=3]{fig/supp_sim.pdf}
\par\vspace{-0.5em}
\caption{Transfer of \method results on \bench to dexterous robot hands in Genesis (continued).}
\label{fig:app_sim_3}}
\vspace*{\fill}
\newpage

\section{Experimental Details}
\label{app:details}
This section extends the experimental setup of the main paper with details of \method and \bench.
We first describe the implementation in \Cref{app:impl}.
We then present the solvers for each kind of matter in \Cref{app:sim}.
Next, we summarize the reconstruction clips and generation tasks in \Cref{app:bench}.
Finally, we detail the evaluation protocol and metrics in \Cref{app:metrics}.

\subsection{Implementation Details}
\label{app:impl}\label{app:env}\label{app:agents}
The two generators of Eqn.~\eqref{eq:loop} carry their context across rounds differently.
The Object Agent keeps one context, whereas each HOI Agent round restarts from the earlier code, reviews and notes so that it does not replay the long simulation sessions of earlier rounds.
The critics do not edit code but judge measurements with Blender~4.5~\cite{blender2025} renders of key frames.
Interaction rounds are ranked lexicographically by acceptance, completeness, unmet plan outcomes, persistent hand faults, major review issues, other checker failures and critic scores.
This ranking uses no ground truth of \bench, and each revision starts from the best round so far.
Hashing each run's copy of the framework reveals any edit of the tools or checker.

Within each round, the agents build on interchangeable off-the-shelf models that ground their code in measured geometry.
A~depth model~\cite{lin2026depthanything3} gives the Object Agent the camera and metric depth of the input, a segmentation model~\cite{carion2026sam3} gives object and hand masks and a hand reconstruction model~\cite{potamias2025wilor} gives real-size hand meshes.
The object critic sees input overlays, eight-view turntables and renders at joint limits.
The HOI Agent sets 21 keypoints per hand by the wrist and the flexion, abduction and twist of each finger.
MANO~\cite{romero2017mano} is fitted to them by Levenberg-Marquardt~\cite{levenberg1944method,marquardt1963algorithm} under joint limits, temporal smoothness and its pose prior, weighting contact keypoints higher.
As plans rarely time every reach and carry, Fitts' law~\cite{fitts1954information} fitted to five hand datasets~\cite{chao2021dexycb,zhan2024oakink2,fan2023arctic,banerjee2025hot3d,lim2026hrdexdb} gives each untimed one a human duration from its distance.

Reconstruction keeps these tools but follows the video instead of a plan, posing or simulating every object in every frame without a pose tracker.
Depth and masks guide the objects, while pose-estimator keypoints~\cite{khirodkar2026sapiens2} and hand meshes initialize the hands.
A fixed camera keeps its first-frame pose, whereas egocentric clips take per-frame poses from the depth model.
After every build, the agent compares its overlay with the video by mask overlap, keypoint distance and depth offset.
A real-size hand that matches the keypoints fits at only one depth, so the objects are scaled about the camera until the hands touch them.

\subsection{Physics Simulation}
\label{app:sim}
Each non-rigid object that the HOI Agent simulates runs in the solver for its kind of matter $\kappa^n$ of Eqn.~\eqref{eq:object}, and the agent may tune its parameters $\boldsymbol\xi^n$, resolution and time step.
The four solvers below cover matter that loses its shape, elastic solids, liquids and sheets or cords.
Each hand enters every solver as a closed MANO surface whose winding number~\cite{barill2018fastwinding} marks its inside.
\begin{enumerate}[leftmargin=*, noitemsep, topsep=-\parskip, label=\roman*.]
  \item \textbf{MPM}~\cite{hu2018mlsmpm} handles matter that loses its rest shape, as its particles need no mesh and carry the deformation history.
  Its elastic law is the fixed-corotated energy~\cite{stomakhin2013snow} $\psi(\mathbf F)=\mu\|\mathbf F-\mathbf R\|_F^2+\frac{\lambda}{2}(J-1)^2$ of the deformation gradient $\mathbf F$, where $\mathbf R$ is the rotation of $\mathbf F$, $J=\det\mathbf F$ and $\mu$ and $\lambda$ are the Lam\'e parameters.
  Plastic and granular~\cite{klar2016drucker} laws then project $\mathbf F$ back onto a von Mises or Drucker--Prager yield surface, and viscous matter keeps only its volume change.
  Materials sharing a grid interact both ways (\eg, stirred paints), and particles may carry extra states such as absorbed liquid.
  The step $\Delta t\propto h/c$ scales with the grid spacing $h$ over the wave speed $c=\sqrt{(\lambda+2\mu)/\rho}$ of the stiffest material with density $\rho$.
  \item \textbf{FEM}~\cite{sifakis2012fem} handles elastic solids that return to their rest mesh.
  Embedding the mesh in tetrahedra makes the rendered surface the simulated one.
  Each tetrahedron computes $\mathbf F$ from its deformed and rest edges and enforces the two terms of $\psi$ as XPBD constraints (iv) with compliances $1/(2\mu V)$ and $1/(\lambda V)$ for its rest volume $V$.
  \item \textbf{FLIP}~\cite{zhu2005sand} handles liquids that pour and splash.
  Each step makes the grid velocity divergence-free by solving a Poisson equation for the pressure on a staggered grid.
  Each particle velocity $\mathbf v_p$ then blends the grid velocity $\mathbf v_g$ and its change $\Delta\mathbf v_g$ as ${\mathbf v_p\leftarrow0.95\,(\mathbf v_p+\Delta\mathbf v_g)+0.05\,\mathbf v_g}$ so that splashes are not smoothed away.
  \item \textbf{XPBD}~\cite{macklin2016xpbd,macklin2019smallsteps} handles cloth, paper and rope on their meshes, as they are thinner than a grid cell.
  It projects each constraint $C_j(\mathbf x)=0$ on the point positions $\mathbf x$ with a compliance $\alpha_j$ (\ie, inverse stiffness) so that the stiffness does not depend on the time step or iteration count.
  Stretch constraints come with implicit bending~\cite{bergou2006bending} and self-collision, and plastic bending, wetting and tearing are optional.
  The step shrinks with the hand speed so that a sheet does not tunnel through a finger, and a pinch holds only by friction unless the program attaches the sheet.
\end{enumerate}
Rigid and articulated objects instead use Bullet~\cite{coumans2015bullet} in Blender, which keeps an object dynamic until a grasp closes, kinematic while held and dynamic again after release.
The grip check of \Cref{app:grip} then tests whether the contacts could hold it.
Coupling across solvers is one-way, as the frames of one simulation enter another as colliders, sources, sinks or a surrounding liquid (\eg, cereal in MPM floats on milk from FLIP without pushing it back).
For scoring, liquids are submitted as closed particle surfaces, grains as small ellipsoids, solids as their embedded surfaces and sheets and cords as thickened shells and tubes.

\subsection{Benchmark Details}
\label{app:bench}
\label{app:bench:video_gt}
\providecommand{\benchTasks}{80}\providecommand{\benchScene}{40}\providecommand{\benchText}{20}\providecommand{\benchImage}{20}\providecommand{\benchScenes}{77}\providecommand{\benchMetrics}{38}\providecommand{\benchRareMetrics}{9}%
We detail how the 20 reconstruction clips and \benchTasks{} generation tasks of \bench were selected and annotated.
The clips were fixed before any experiment and cover each kind of matter with five clips.
Rigid clips come from ARCTIC~\cite{fan2023arctic} and DexYCB~\cite{chao2021dexycb}, deformable ones from PhysCoRe~\cite{yin2026physcore} and GigaHands~\cite{fu2025gigahands}, co-dimensional ones from PhysTwin~\cite{jiang2025phystwin} and flowing ones from OakInk2~\cite{zhan2024oakink2} and H2O~\cite{kwon2021h2o}.
The rigid clips show distinct objects and keep the hands in view and in contact for most of each clip.
Clips from ARCTIC, OakInk2 and H2O are egocentric, whereas the others use a fixed camera.
Each clip is paired with a sentence that names the interaction to reconstruct.

To score these clips, we take the hand and object annotations released with each dataset.
Two cases need new MANO fits.
In two GigaHands clips the released right-hand fits exceed the joint limits, so we refit MANO to the multi-view joints under joint limits and a pose prior.
PhysTwin and PhysCoRe provide three calibrated RGB-D views and hand masks but no MANO hands.
Following multi-view hand annotation~\cite{chao2021dexycb,hampali2020honnotate,kwon2021h2o,wang2024hocap,lee2026physhandi}, we fit MANO to triangulated Sapiens2~\cite{khirodkar2026sapiens2} keypoints and depth under joint limits, a pose prior and an acceleration penalty.
Object annotations vary more across datasets.
ARCTIC, DexYCB, OakInk2 and H2O release posed object meshes that include the articulation in ARCTIC but cover only the containers of flowing matter.
PhysTwin and PhysCoRe annotate objects as tracked depth points on the surfaces the camera sees, and GigaHands annotates no object.
No dataset annotates flowing matter in 3D, so it is scored by containment.

\label{app:bench:generation}
Generation tasks have no reference motion and are instead scored by hidden success conditions that the agents do not see.
Each task starts from a 3D scene, a photograph or text alone.
A 3D scene supplies metric object meshes from YCB scans, ARCTIC objects, 4DCodeBench worlds~\cite{shen2026fourdcodebench} and procedural geometry.
A photograph is a stock image without hands that comes with named object boxes, whereas a text task leaves the whole scene to the agent.
The conditions check placement, orientation, release, stability, deformation, containment and material-specific changes.
A part named in a condition (\eg, a handle) follows fixed geometric rules in a supplied scene and is otherwise located on the objects the agent submits.
\Cref{app:catalogue} lists every task with its hidden annotations.

\subsection{Evaluation Protocol and Metrics}
\label{app:metrics}\label{app:metrics:video}
Reconstruction measures how closely the hands, objects and non-rigid matter follow the clip, whereas generation measures task success, physical plausibility and perceived quality.
Reconstructions are scored frame by frame in the camera coordinates of the input without alignment unless stated below, so errors in metric scale and depth count in full.
Lengths are in millimeters, Vol.\ and In in percent and Acc.\ in $\mathrm m/\mathrm s^2$, while C-F1 and both IoUs range from 0 to 1.

\vspace{0.5em}\noindent\textbf{Hands.}\quad
MJE averages the wrist-relative error of the 21 keypoint positions $\mathbf J_j$ of a hand with the wrist at $j=0$,
\begin{equation}
  \mathrm{MJE}=\frac{1}{|\mathcal J|}\sum_{j\in\mathcal J}\bigl\|(\hat{\mathbf J}_j-\hat{\mathbf J}_0)-(\mathbf J_j-\mathbf J_0)\bigr\|_2,
\end{equation}
where $\mathcal J$ indexes the keypoints and hats mark predictions.
It is averaged over all frames and hands.
Acc.\ applies the same error to the accelerations of the wrist-relative MANO vertices, computed by second differences over frames~\cite{fan2023arctic}.

\vspace{0.5em}\noindent\textbf{Objects and contact.}\quad
Object surfaces are compared by the Chamfer distance between their point samples $\hat{\mathcal P}$ and $\mathcal P$,
\begin{equation}
  \mathrm{CD}(\hat{\mathcal P},\mathcal P)=\frac{1}{|\hat{\mathcal P}|}\sum_{\hat{\mathbf p}\in\hat{\mathcal P}}\min_{\mathbf p\in\mathcal P}\|\hat{\mathbf p}-\mathbf p\|_2+\frac{1}{|\mathcal P|}\sum_{\mathbf p\in\mathcal P}\min_{\hat{\mathbf p}\in\hat{\mathcal P}}\|\mathbf p-\hat{\mathbf p}\|_2.
\end{equation}
Following \citet{xie2026cari4d}, CD$_o$ averages this distance over frames after one similarity transform fitted to the first-frame hands.
MRRPE~\cite{fan2023arctic} is the error of the object center relative to the wrist.
C-F1 averages the F1 score $2|\hat C\cap C|/(|\hat C|+|C|)$ over the frames with contact on either side, where $\hat C$ and $C$ hold the MANO vertices within \SI{10}{\milli\meter} of or inside the object.
Flowing clips score CD$_o$ and C-F1 on their containers, as the matter itself has no GT.
Pen.\ averages over frames the deepest penetration between a hand and an object.

\vspace{0.5em}\noindent\textbf{Non-rigid matter.}\quad
Following \citet{jiang2025phystwin}, CD keeps only the one-sided term from the tracked points to the predicted surface under the L1 norm.
Track follows the predicted vertex nearest to each tracked point at the first frame and measures its error over the clip.
IoU$_\mathrm h$ compares object silhouettes in held-out camera views with hand pixels excluded, and IoU$_\mathrm v$ compares them with the masks of the input video.
PhysTwin and PhysCoRe use dataset masks, and the other clips use SAM~3~\cite{carion2026sam3} masks that the agents also receive.
Vol.\ is the largest relative volume change of a closed deformable mesh.
In averages the share of matter inside the source container at the start and inside the receiving container at the end of the clip.

\vspace{0.5em}\noindent\textbf{Task success.}\quad
\label{app:metrics:scoring}
For the hidden conditions $\mathcal C$ of a task, the success score is
\begin{equation}
  S=\frac{1}{|\mathcal C|}\sum_{c\in\mathcal C}\mathbbm{1}\bigl[c(P)\bigr],
\end{equation}
where the indicator $\mathbbm{1}[c(P)]$ is one when $c$ holds in the output of the submitted program $P$ of Eqn.~\eqref{eq:task}.
A condition that cannot be measured (\eg, because its object or part is missing) counts as unmet.
Meeting every condition ($S=1$) is necessary but not sufficient to solve a task.
The result must also follow the part roles, use the hands the task names, pass every validity check and meet every physical threshold in \Cref{tab:cat_checks}.
SR is the percentage of solved tasks and involves no model judgment.

\vspace{0.5em}\noindent\textbf{Physical plausibility.}\quad
Whereas SR only thresholds the physical checks of \Cref{tab:cat_checks}, Phys grades them to measure how physically plausible a result is.
Each check maps its error $e$ to the score
\begin{equation}
  s=\mathrm{clip}\Bigl(\frac{e_0-e}{e_0-e_1},\,0,\,1\Bigr),
\end{equation}
which falls linearly from one at the full-credit error $e_1$ to zero at the zero-credit error $e_0$.
Only the check on cloth or rope pinned away from the hands is pass or fail.
Phys takes the lowest score among the checks that apply to the result, so a single severe violation cannot be offset by the other checks.
\label{app:grip}Since the hands are kinematic, the grip check asks whether forces $\mathbf f_k$ at the $K$ measured contacts $\mathbf p_k$ could hold a carried object.
With linearized friction cones it solves the grasp wrench problem~\cite{ferrari1992planning}
\begin{equation}
  \min_{\mathbf f_1,\dots,\mathbf f_K}\ \sum_{k=1}^{K}f_k^{\perp}
  \quad\text{s.t.}\quad
  \sum_{k=1}^{K}\begin{bmatrix}\mathbf f_k\\(\mathbf p_k-\mathbf c)\times\mathbf f_k\end{bmatrix}
  =\begin{bmatrix}m(\mathbf a-\mathbf g)\\\mathbf I\dot{\boldsymbol\omega}+\boldsymbol\omega\times\mathbf I\boldsymbol\omega\end{bmatrix},
  \quad\|\mathbf f_k^{\parallel}\|\leq\mu f_k^{\perp}\ \ \forall k,
\end{equation}
where $f_k^{\perp}$ and $\mathbf f_k^{\parallel}$ are the normal and tangential force components and $m$, $\mathbf c$, $\mathbf I$, $\mathbf a$ and $\boldsymbol\omega$ are the mass, center of mass, inertia, acceleration and angular velocity of the object under gravity $\mathbf g$.
A frame passes when the minimal total normal force is at most ten times the weight, where the friction coefficient $\mu$ is measured for skin on each material~\cite{buchholz1988effects,tomlinson2011understanding,seo2009friction,derler2009friction}.

\vspace{0.5em}\noindent\textbf{Perceptual judge.}\quad
\label{app:metrics:judge}
SR and Phys involve no model, whereas the perceived quality of each result is assessed by a single judge model.
The judge sees the instruction and the submitted result but neither the plan nor the checker report.
It reviews the same renders as the interaction critic and scores hand naturalness, contact, object appearance and motion from 1 to 5.
It also picks the better result of each pair, and a pair judged differently in the two orders counts as a tie worth half a win to each side.
Bradley-Terry strengths $\pi_i$ of backbone $i$ fitted by the MM algorithm~\cite{hunter2004mm} model these preferences and give the Elo ratings,
\begin{equation}
  P(i\succ j)=\frac{\pi_i}{\pi_i+\pi_j},\qquad
  \mathrm{Elo}_i=1000+400\log_{10}\frac{\pi_i}{\bar\pi},
\end{equation}
where $\bar\pi$ is the geometric mean of all strengths.
A program that fails to execute loses all its pairs unjudged.

\vspace{0.5em}\noindent\textbf{Part roles and validity.}\quad
\label{app:metrics:families}
A solved task must also respect the part roles and the named hands and pass every validity check.
The hands must touch every part they are meant to use and keep clear of every part that is unsafe or kept free, whereas touching a part to avoid is only penalized.
A task that names its hands requires exactly those hands to touch the targets.
Missing hands or targets and hands that never reach a target score zero, so avoiding an object cannot win on penetration.
Validity checks cover the files, object sizes, hands and provenance of non-rigid matter.
Hand checks require the supplied hand shape and flag implausible poses and motion (\eg, joints past their limits or interpenetrating hands).
For provenance, the scorer reruns the saved simulation of every non-rigid object and fails any that no solver moved or that departs from its rerun.
During a run, the checker applies the hand, role and physical checks with the plan replacing the hidden conditions.

\section{Additional Results}
\label{app:analyses}
This section extends the experiments of the main paper with additional quantitative results of \method.
We first analyze the gain and cost of revision with and without feedback in \Cref{app:cost} and show in \Cref{app:nophys} how objects generated without our physics simulation can break when transferred to simulation.
We then report bootstrap intervals over tasks in \Cref{app:stats} and test whether the results depend on the judge in \Cref{app:judge} and on the solved thresholds in \Cref{app:sensitivity}.
Finally, we break down the generation results by input in \Cref{app:gen_inputs} and analyze the hand error in reconstruction in \Cref{app:recon_hands}.

\clearpage
\subsection{Revision Efficiency and Feedback}
\label{app:cost}
Revising a result in the closed loop of Eqn.~\eqref{eq:loop} takes extra rounds, so we measure what they gain and cost.
\Cref{tab:app:revision} details the results behind \Cref{fig:analysis:cost} in terms of success, acceptance and efficiency.
Every run stores each round, so round~0 gives a baseline with the checker but without critique at no extra cost.
At each revision budget we score the round that the loop would retain by its own ranking of \Cref{app:impl}, and runs that stop early keep their final result.
We test the gain for significance with an exact McNemar test~\cite{mcnemar1947note}, a paired test on the tasks that only one of round~0 and the retained result solves.
Revision gains 11 tasks and loses none for both Opus~5.5 and GPT-6 Astra, which is significant with $p=0.001$.
Sonnet~5.5 and GPT-6.1 Sol each gain 10 tasks but also lose 5, as the loop ranks its rounds without the hidden annotations.
The Evaluator Agent accepts 33.8 to 71.3\% of the runs, and acceptance reliably signals success since 80.7 to 91.3\% of the accepted results are solved.
Rejection is stricter than SR, as 16.1 to 50.0\% of the rejected results are also solved.
Acceptance also requires the approval of the critic and the goals of the agent's own plan, neither of which enters SR.
A run averages 1.49 to 2.33 revisions, 145.1 to 181.4 thousand output tokens over all agent calls and \$4.79 to \$43.67 at standard API rates.
Cost thus depends more on the price of each model than on its tokens, as GPT-6 Astra writes the fewest tokens yet costs the most.

\begin{table}[h]
  \centering
  \caption{Revision efficiency on the 80 generation tasks. SR$_0$ scores round~0 (\ie, written with the checker before any critique) and SR the result that the loop retains after at most three revisions. $\Delta$SR is their difference with its 95\% bootstrap interval and exact McNemar~\cite{mcnemar1947note} $p$-value. Accept is the share of runs that the Evaluator Agent accepts, Prec.\ the share of accepted results that are solved and Rej.\ solved the share of rejected results that are solved, all in percent. Revisions, Tokens (K) and Cost (\$) at standard API rates are means per run.}
  \label{tab:app:revision}
  \small
  \resizebox{\linewidth}{!}{%
  \begin{tabular}{l|ccN{\mathbf{+13.8}}{9.4}c|ccc|ccc}
    \toprule
    & \multicolumn{4}{c|}{\textbf{Success}} & \multicolumn{3}{c|}{\textbf{Acceptance}} & \multicolumn{3}{c}{\textbf{Efficiency}} \\
    \cmidrule(lr){2-5}\cmidrule(lr){6-8}\cmidrule(lr){9-11}
    \multicolumn{1}{c|}{\textbf{Model}} & SR$_0$ ($\uparrow$) & SR ($\uparrow$) & \multicolumn{1}{c}{$\Delta$SR ($\uparrow$)} & $p$ & Accept & Prec.\ ($\uparrow$) & Rej.\ solved ($\downarrow$) & Revisions ($\downarrow$) & Tokens ($\downarrow$) & Cost ($\downarrow$) \\
    \midrule
    \modelicon{claude}Claude Opus 5.5 & 47.5 & 61.3 & \alignednum{$\mathbf{+13.8}$}[7.7] & 0.001 & 61.3 & \underline{89.8} & \textbf{16.1} & 1.73 & \underline{164.9} & \phantom{0}9.18 \\
    \modelicon{claude}Claude Sonnet 5.5 & 45.0 & 51.3 & \alignednum{$\underline{+6.3}$}[9.3] & 0.302 & 33.8 & 81.5 & 35.8 & 2.33 & 170.9 & \phantom{0}\textbf{4.79} \\
    \modelicon{chatgpt}GPT-6 Astra & \textbf{60.0} & \textbf{73.8} & \alignednum{$\mathbf{+13.8}$}[7.6] & 0.001 & 57.5 & \textbf{91.3} & 50.0 & \underline{1.66} & \textbf{145.1} & 43.67 \\
    \modelicon{chatgpt}GPT-6.1 Sol & \underline{57.5} & \underline{63.8} & \alignednum{$\underline{+6.3}$}[9.4] & 0.302 & 71.3 & 80.7 & \underline{21.7} & \textbf{1.49} & 181.4 & \phantom{0}\underline{6.80} \\
    \bottomrule
  \end{tabular}}
\end{table}

\begin{wraptable}[10]{r}{0.54\linewidth}
\centering\small
\vspace{-13pt}
\caption{Revision with feedback against fresh attempts for Opus~5.5 on the 40 tasks whose round~0 is rejected and unsolved. Tokens (K) and Cost (\$) are cumulative means per run. $\Delta$SR is the gain of feedback, and $n_1$/$n_2$ count the tasks that only feedback or only fresh attempts solve.}
\label{tab:app:nofeedback}
  \setlength{\tabcolsep}{2.5pt}
  \resizebox{\linewidth}{!}{%
  \begin{tabular}{c|ccc|ccc|N{+7.5}{17.1}c}
    \toprule
    & \multicolumn{3}{c|}{\textbf{With Feedback}} & \multicolumn{3}{c|}{\textbf{Fresh Attempts}} & \multicolumn{2}{c}{\textbf{Paired}} \\
    \cmidrule(lr){2-4}\cmidrule(lr){5-7}\cmidrule(lr){8-9}
    \textbf{Revisions} & SR ($\uparrow$) & Tokens ($\downarrow$) & Cost ($\downarrow$) & SR ($\uparrow$) & Tokens ($\downarrow$) & Cost ($\downarrow$) & \multicolumn{1}{c}{$\Delta$SR} & $n_1$/$n_2$ \\
    \midrule
    0 & \phantom{0}0.0 & 102.3 & \phantom{0}5.63 & \phantom{0}0.0 & 102.3 & \phantom{0}5.63 & \alignednum{$0.0$} & 0/0 \\
    1 & 10.0 & 144.0 & \phantom{0}8.02 & 10.0 & 165.0 & \phantom{0}9.76 & \alignednum{$0.0$}[12.0] & 3/3 \\
    2 & 22.5 & 176.4 & 10.00 & 15.0 & 223.8 & 13.48 & \alignednum{$+7.5$}[14.5] & 6/3 \\
    3 & 27.5 & 199.6 & 11.45 & 22.5 & 279.9 & 16.95 & \alignednum{$+5.0$}[17.1] & 7/5 \\
    \bottomrule
  \end{tabular}}
\vspace{-1em}
\end{wraptable}
\vspace{0.5em}\noindent\textbf{Revision without feedback.}\quad
\label{app:nofeedback}
These gains could still come from retrying rather than from feedback, since every revision is also one more attempt.
We separate the two with a control that keeps the loop but starts every round after round~0 anew from the task and its inputs, without the code, reviews or notes of earlier rounds.
The checker, the critic, the ranking of rounds and the stop at acceptance stay unchanged.
The control does not stop early for lack of improvement, so it makes at least as many attempts as the loop.
We run it with Opus~5.5 on the 40 generation tasks whose round~0 is rejected and unsolved, which include all 11 tasks that revision solves.
\Cref{tab:app:nofeedback} shows that feedback matches or exceeds the control at every revision budget while spending fewer tokens.
It solves 9 of these tasks after two revisions, which the control reaches only after three with 1.6 times the output tokens and 1.7 times the cost.
Its acceptance is also more reliable, as 9 of the 12 results accepted with feedback are solved against 3 of the 7 in the control.
Feedback thus makes revision both cheaper and more reliable than plain retrying.

\subsection{Transfer Without Physics Simulation}
\label{app:nophys}
\begin{figure}[t]
\centering
\includegraphics[width=\linewidth]{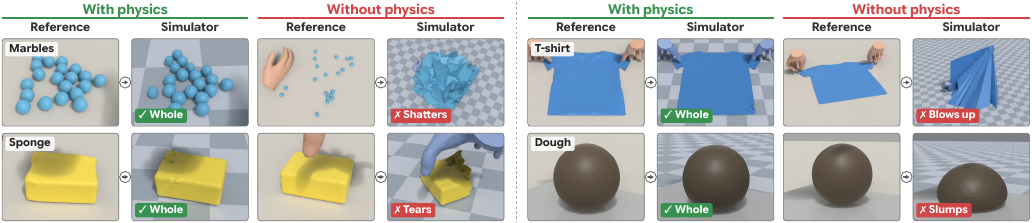}
\vspace{-1.5em}
\caption{Objects generated with and without our physics simulation and transferred to Genesis~\cite{genesis2024} with MANO hands. Each pair shows the same moment as written (Reference) and in Genesis (Simulator), and each tag marks whether the object stays whole or how it breaks.}
\label{fig:app_nophys}
\end{figure}

\Cref{sec:exp:sim} transfers our results to Genesis~\cite{genesis2024}, where each object responds according to the physics that it carries.
We test what our four solvers contribute by regenerating tasks with Opus~5.5 without them and their material presets.
The HOI Agent then moves each non-rigid object with its own code (\eg, a simplified simulator or a drawn deformation) or with the built-in physics of Blender.
Genesis simulates each object from its stated matter and parameters with the same settings in both versions, and MANO hands replace the robot so that any failure comes from the objects.
\Cref{fig:app_nophys} shows four objects that stay whole with the solvers but break without them.
The marbles burst into shards and the T-shirt blows up before the hands reach it.
The sponge is squeezed as written but hardly yields in Genesis and tears when put down.
The dough is stated too soft to bear its own weight, so it slumps before the hand arrives.
Without our solvers the written motion need not follow the stated material.
Our solvers instead derive it from this material, which helps the result transfer to another engine.

\subsection{Statistical Uncertainty}
\label{app:stats}
The generation results of \Cref{tab:gen} come from one run of each backbone per task, so they vary with both the sampled tasks and the randomness of each run.
Resampling the tasks with replacement~\cite{efron1979bootstrap} covers both, since a drawn task brings the outcome of its run and all its comparisons.
We report 95\% intervals as 1.96 bootstrap standard errors on either side of each result over 10,000 resamples for SR and 1,000 for Elo.
\Cref{tab:app:ci} shows that the 80 tasks determine the SR of a backbone to within about 10 points and the 20 tasks of one kind of matter to within 13 to 22 points.
The SR intervals of GPT-6 Astra and Sonnet~5.5 do not overlap, so the 80 tasks separate the strongest backbone from the weakest.
For Elo the intervals of both GPT backbones lie above those of both Claude backbones.
The gap in visual quality between the model families thus exceeds this uncertainty.
\begin{table}[h]
\centering
\caption{SR and Elo of \Cref{tab:gen} with 95\% intervals from resampling the tasks. Each kind of matter holds 20 tasks and All the 80 tasks.}
\label{tab:app:ci}\small
  \setlength{\tabcolsep}{6pt}
  \resizebox{\linewidth}{!}{%
  \begin{tabular}{c|l|N{1102.0}{110.4}N{1102.0}{110.4}N{1102.0}{110.4}N{1102.0}{110.4}|N{1102.0}{46.2}}
    \toprule
    \textbf{Metric} & \multicolumn{1}{c|}{\textbf{Model}} & \multicolumn{1}{c}{\textcolor{matRigid}{\textbf{Rigid}}} & \multicolumn{1}{c}{\textcolor{matDeform}{\textbf{Deformable}}} & \multicolumn{1}{c}{\textcolor{matCodim}{\textbf{Co-dimensional}}} & \multicolumn{1}{c|}{\textcolor{matFlow}{\textbf{Flowing}}} & \multicolumn{1}{c}{\textbf{All}} \\
    \midrule
    \multirow{4}{*}{SR ($\uparrow$)} & \modelicon{claude}Claude Opus 5.5 & \alignednum{$70.0$}[20.3] & \alignednum{$55.0$}[21.7] & \alignednum{$60.0$}[21.4] & \alignednum{$60.0$}[21.3] & \alignednum{$61.3$}[10.7] \\
     & \modelicon{claude}Claude Sonnet 5.5 & \alignednum{$65.0$}[20.9] & \alignednum{$40.0$}[21.3] & \alignednum{$50.0$}[22.1] & \alignednum{$50.0$}[21.8] & \alignednum{$51.3$}[11.0] \\
     & \modelicon{chatgpt}GPT-6 Astra & \alignednum{$90.0$}[13.3] & \alignednum{$70.0$}[19.9] & \alignednum{$55.0$}[21.8] & \alignednum{$80.0$}[17.4] & \alignednum{$73.8$}[9.6] \\
     & \modelicon{chatgpt}GPT-6.1 Sol & \alignednum{$85.0$}[15.8] & \alignednum{$55.0$}[21.9] & \alignednum{$55.0$}[21.8] & \alignednum{$60.0$}[21.5] & \alignednum{$63.8$}[10.5] \\
    \midrule
    \multirow{4}{*}{Elo~\cite{elo1978rating} ($\uparrow$)} & \modelicon{claude}Claude Opus 5.5 & \alignednum{$963.9$}[79.3] & \alignednum{$951.0$}[84.8] & \alignednum{$982.8$}[60.5] & \alignednum{$920.9$}[77.5] & \alignednum{$954.5$}[38.1] \\
     & \modelicon{claude}Claude Sonnet 5.5 & \alignednum{$888.5$}[99.4] & \alignednum{$923.6$}[77.6] & \alignednum{$874.0$}[101.5] & \alignednum{$896.6$}[110.4] & \alignednum{$895.2$}[46.2] \\
     & \modelicon{chatgpt}GPT-6 Astra & \alignednum{$1102.0$}[76.5] & \alignednum{$1058.1$}[87.8] & \alignednum{$1069.3$}[76.3] & \alignednum{$1098.5$}[59.1] & \alignednum{$1082.3$}[37.2] \\
     & \modelicon{chatgpt}GPT-6.1 Sol & \alignednum{$1045.6$}[87.0] & \alignednum{$1067.3$}[83.9] & \alignednum{$1074.0$}[88.7] & \alignednum{$1084.0$}[78.3] & \alignednum{$1068.1$}[42.9] \\
    \bottomrule
  \end{tabular}}
\end{table}

\suppneedspace{12\baselineskip}
\subsection{Judge Consistency}
\label{app:judge}
\begin{wraptable}[10]{r}{0.47\linewidth}
\centering\small
\vspace{-13.9pt}
\caption{Single judgments on the 120 pairs over the 20 tasks of \Cref{tab:analysis:reliability}. Order Changes is the percentage of pairs whose preference flips when the order is reversed. Above the diagonal is the agreement of two judges and below it their Cohen's $\kappa$~\cite{cohen1960kappa}.}
\label{tab:app:judge}
  \setlength{\tabcolsep}{3pt}
  \resizebox{\linewidth}{!}{%
  \begin{tabular}{l|c|cccc}
    \toprule
    & & \multicolumn{4}{c}{\textbf{Agreement (\%) / $\kappa$}} \\
    \cmidrule(lr){3-6}
    \multicolumn{1}{c|}{\textbf{Judge}} & \textbf{Order Changes} ($\downarrow$) & \modelicon{claude}Opus & \modelicon{claude}Sonnet & \modelicon{chatgpt}Astra & \modelicon{chatgpt}Sol \\
    \midrule
    \modelicon{claude}Claude Opus 5.5 & 20.0 & -- & 69.2 & 67.5 & 70.0 \\
    \modelicon{claude}Claude Sonnet 5.5 & 16.7 & 0.60 & -- & 70.8 & 68.3 \\
    \modelicon{chatgpt}GPT-6 Astra & 11.7 & 0.58 & 0.62 & -- & 84.2 \\
    \modelicon{chatgpt}GPT-6.1 Sol & 15.8 & 0.61 & 0.59 & 0.79 & -- \\
    \bottomrule
  \end{tabular}}
\vspace{-1em}
\end{wraptable}
Elo also depends on the judge model, which \Cref{tab:analysis:reliability} examines with the ratings of four judges.
Here we analyze the single judgments behind these ratings with two statistics in \Cref{tab:app:judge}.
We hold the generated results fixed and let each backbone at high reasoning effort judge the same 20 generation tasks, which cover five tasks per kind of matter with the inputs in the proportion of the benchmark.
Every judge compares every pair of backbones on each task with the same renders and instructions in both presentation orders.
The first statistic is how often a preference changes when the order is reversed, which turns the pair into a tie.
The second is how often two judges prefer the same result, with Cohen's $\kappa$~\cite{cohen1960kappa} correcting for agreement by chance.
The order changes the preference in 11.7 to 20.0\% of the pairs, so judging both orders keeps this bias from favoring either result.
Any two judges agree on 67.5 to 84.2\% of the pairs with $\kappa$ of 0.58 to 0.79, well above chance.
Opus~5.5 agrees with the two GPT judges about as often as with Sonnet~5.5, so the main judge agrees across families as well as within its own.
Elo aggregates many such judgments, so the order of the families holds under every judge in \Cref{tab:analysis:reliability}.

\subsection{Sensitivity to Solved Thresholds}
\label{app:sensitivity}
Unlike Elo~\cite{elo1978rating}, SR involves no judge but rests on the thresholds of the physical checks in \Cref{tab:cat_checks}, which decide whether a task is solved.
We test this dependence by varying one threshold at a time and rescoring the 80 generation tasks.
A last setting also requires the judge to rate hand naturalness, contact, object appearance and motion each at least 3 on its scale from 1 to 5.
\Cref{tab:app:sensitivity} shows that SR changes little with these thresholds, although the checks do filter results.
Tightening the penetration bound lowers SR by up to 3.8, and the other bounds change it by at most 1.3.
Loosening the bounds on points inside a hand and on unsupported motion leaves SR unchanged, so unsolved tasks fail for reasons other than these bounds.
Requiring the judge lowers SR by 7.5 to 22.5 because success then also demands visual quality, which Elo measures separately.
The order of the backbones stays the same in every setting, so our comparison does not rest on the thresholds.
\begin{table}[h]
\centering
\caption{SR on the 80 generation tasks under varied solved thresholds. Pen.\ bounds the penetration, Points the share of points inside a hand and Unsup.\ the unsupported object motion. Judge is the score from 1 to 5 that each criterion of the judge must reach where required. The first row is the default of \Cref{tab:cat_checks} and every other row changes one setting. $\Delta$ is the change from the default, shown in \textcolor{blue!80!black}{blue} when SR falls.}
\label{tab:app:sensitivity}\small
  \setlength{\tabcolsep}{5pt}
  \resizebox{\linewidth}{!}{%
  \begin{tabular}{cccc|cc|cc|cc|cc|cc}
    \toprule
    \multicolumn{4}{c|}{\textbf{Setting}} & \multicolumn{2}{c|}{\modelicon{claude}\textbf{Claude Opus 5.5}} & \multicolumn{2}{c|}{\modelicon{claude}\textbf{Claude Sonnet 5.5}} & \multicolumn{2}{c|}{\modelicon{chatgpt}\textbf{GPT-6 Astra}} & \multicolumn{2}{c|}{\modelicon{chatgpt}\textbf{GPT-6.1 Sol}} & \multicolumn{2}{c}{\textbf{Average}} \\
    \cmidrule(lr){1-4}\cmidrule(lr){5-6}\cmidrule(lr){7-8}\cmidrule(lr){9-10}\cmidrule(lr){11-12}\cmidrule(lr){13-14}
    Pen. & Points & Unsup. & Judge & SR ($\uparrow$) & $\Delta$ & SR ($\uparrow$) & $\Delta$ & SR ($\uparrow$) & $\Delta$ & SR ($\uparrow$) & $\Delta$ & SR ($\uparrow$) & $\Delta$ \\
    \midrule
    \rowcolor{tabgray}5\,mm & 5\,\% & 10\,\% & -- & 61.3 & -- & 51.3 & -- & 73.8 & -- & 63.8 & -- & 62.5 & -- \\
    3\,mm & 5\,\% & 10\,\% & -- & 57.5 & \phantom{0}\textcolor{blue!80!black}{$-$3.8} & 47.5 & \phantom{0}\textcolor{blue!80!black}{$-$3.8} & 72.5 & \textcolor{blue!80!black}{$-$1.3} & 63.8 & \phantom{$+$}0.0 & 60.3 & \phantom{0}\textcolor{blue!80!black}{$-$2.2} \\
    5\,mm & 2\,\% & 10\,\% & -- & 61.3 & \phantom{0}\phantom{$+$}0.0 & 51.3 & \phantom{0}\phantom{$+$}0.0 & 73.8 & \phantom{$+$}0.0 & 63.8 & \phantom{$+$}0.0 & 62.5 & \phantom{0}\phantom{$+$}0.0 \\
    5\,mm & 10\,\% & 10\,\% & -- & 61.3 & \phantom{0}\phantom{$+$}0.0 & 51.3 & \phantom{0}\phantom{$+$}0.0 & 73.8 & \phantom{$+$}0.0 & 63.8 & \phantom{$+$}0.0 & 62.5 & \phantom{0}\phantom{$+$}0.0 \\
    5\,mm & 5\,\% & 5\,\% & -- & 61.3 & \phantom{0}\phantom{$+$}0.0 & 51.3 & \phantom{0}\phantom{$+$}0.0 & 72.5 & \textcolor{blue!80!black}{$-$1.3} & 63.8 & \phantom{$+$}0.0 & 62.2 & \phantom{0}\textcolor{blue!80!black}{$-$0.3} \\
    5\,mm & 5\,\% & 20\,\% & -- & 61.3 & \phantom{0}\phantom{$+$}0.0 & 51.3 & \phantom{0}\phantom{$+$}0.0 & 73.8 & \phantom{$+$}0.0 & 63.8 & \phantom{$+$}0.0 & 62.5 & \phantom{0}\phantom{$+$}0.0 \\
    5\,mm & 5\,\% & 10\,\% & $\geq3$ & 47.5 & \textcolor{blue!80!black}{$-$13.8} & 28.8 & \textcolor{blue!80!black}{$-$22.5} & 65.0 & \textcolor{blue!80!black}{$-$8.8} & 56.3 & \textcolor{blue!80!black}{$-$7.5} & 49.4 & \textcolor{blue!80!black}{$-$13.1} \\
    \bottomrule
  \end{tabular}}
\end{table}

\subsection{Breakdown of Generation Results}
\label{app:gen_inputs}
Besides the kind of matter, the tasks of \Cref{tab:gen} differ in their input, which sets how much of the scene the agent must build.
\Cref{tab:gen_inputs} splits the results by input, with 5 tasks per cell for an image or text alone and 10 for a 3D scene.
Non-rigid tasks are hardest from text alone, where the agent must build every object from the instruction.
Their mean SR over the four backbones is 26.7 from text alone against 66.7 from an image and 68.3 from a 3D scene.
Rigid tasks are instead hardest from an image, with a mean SR of 55.0 against 85.0 from either other input.
These tasks demand precise actions on an arrangement that the agent must first recover from the image.
No backbone yet solves the task of pushing one block out of a tower without the tower falling, which leaves headroom for future agents.
Since each input comes with its own tasks, these gaps reflect the tasks as well as the inputs.
The unsolved results also fail for different reasons, as 10 of the 18 rigid ones meet every condition but fail in how the hands act (\eg, a part role or a physical check).
In contrast, 80 of the 102 non-rigid ones leave a condition unmet.
Rigid tasks thus fail more often in the action of the hands and non-rigid tasks in the final state of the matter.
\begin{table}[h]
\centering
\caption{Generation results of \Cref{tab:gen} split by input. Image~+~Text shows the objects in a photograph, Scene~+~Text supplies their meshes and Text~Only leaves the whole scene to the agent. Each cell covers 5 tasks for Image~+~Text and Text~Only and 10 for Scene~+~Text. Within each input the best results are in \textbf{bold} and the second best are \underline{underlined}.}
\label{tab:gen_inputs}\small
  \setlength{\tabcolsep}{4pt}
  \resizebox{\linewidth}{!}{%
  \begin{tabular}{c|l|cccc|cccc|cccc|cccc}
    \toprule
    & & \multicolumn{4}{c|}{\textcolor{matRigid}{\textbf{Rigid}}} & \multicolumn{4}{c|}{\textcolor{matDeform}{\textbf{Deformable}}} & \multicolumn{4}{c|}{\textcolor{matCodim}{\textbf{Co-dimensional}}} & \multicolumn{4}{c}{\textcolor{matFlow}{\textbf{Flowing}}} \\
    \cmidrule(lr){3-6}\cmidrule(lr){7-10}\cmidrule(lr){11-14}\cmidrule(lr){15-18}
    \multicolumn{1}{l|}{\textbf{Input}} & \multicolumn{1}{c|}{\textbf{Model}} & SR ($\uparrow$) & S ($\uparrow$) & Phys ($\uparrow$) & Elo ($\uparrow$) & SR ($\uparrow$) & S ($\uparrow$) & Phys ($\uparrow$) & Elo ($\uparrow$) & SR ($\uparrow$) & S ($\uparrow$) & Phys ($\uparrow$) & Elo ($\uparrow$) & SR ($\uparrow$) & S ($\uparrow$) & Phys ($\uparrow$) & Elo ($\uparrow$) \\
    \midrule
    \multirow{4}{*}{\rotatebox[origin=c]{90}{\begin{tabular}{@{}c@{}}Image\\+ Text\end{tabular}}} & \modelicon{claude}Claude Opus 5.5 & \phantom{0}\underline{40.0} & \underline{0.93} & \textbf{0.90} & \phantom{0}882.5 & \underline{60.0} & \underline{0.80} & \underline{0.99} & \phantom{0}966.3 & \textbf{80.0} & \textbf{0.93} & \textbf{0.80} & \phantom{0}943.6 & \phantom{0}60.0 & \underline{0.93} & 0.75 & 1040.4 \\
     & \modelicon{claude}Claude Sonnet 5.5 & \phantom{0}20.0 & \underline{0.93} & 0.40 & \phantom{0}649.4 & \underline{60.0} & \underline{0.80} & 0.98 & \underline{1033.7} & \underline{60.0} & \underline{0.83} & 0.76 & \phantom{0}867.2 & \phantom{0}40.0 & \underline{0.93} & 0.94 & \phantom{0}805.0 \\
     & \modelicon{chatgpt}GPT-6 Astra & \phantom{0}\textbf{80.0} & \textbf{1.00} & \textbf{0.90} & \textbf{1287.2} & \textbf{80.0} & \textbf{0.93} & 0.94 & \phantom{0}949.3 & \underline{60.0} & \textbf{0.93} & 0.59 & \underline{1016.4} & \textbf{100.0} & \textbf{1.00} & \textbf{1.00} & \underline{1058.7} \\
     & \modelicon{chatgpt}GPT-6.1 Sol & \phantom{0}\textbf{80.0} & \textbf{1.00} & \underline{0.80} & \underline{1180.9} & 40.0 & 0.76 & \textbf{1.00} & \textbf{1050.7} & \textbf{80.0} & \textbf{0.93} & \underline{0.79} & \textbf{1172.8} & \phantom{0}\underline{80.0} & \textbf{1.00} & \underline{0.99} & \textbf{1095.9} \\
    \midrule
    \multirow{4}{*}{\rotatebox[origin=c]{90}{\begin{tabular}{@{}c@{}}Scene\\+ Text\end{tabular}}} & \modelicon{claude}Claude Opus 5.5 & \phantom{0}\textbf{90.0} & 0.93 & \textbf{0.96} & \underline{1025.8} & \underline{60.0} & 0.82 & 0.88 & \phantom{0}917.4 & \textbf{80.0} & 0.83 & \textbf{0.99} & \phantom{0}992.1 & \phantom{0}\underline{80.0} & \textbf{0.97} & \underline{0.99} & \phantom{0}897.0 \\
     & \modelicon{claude}Claude Sonnet 5.5 & \phantom{0}\underline{80.0} & 0.92 & \textbf{0.96} & \phantom{0}948.4 & 40.0 & \underline{0.85} & 0.84 & \phantom{0}907.9 & 60.0 & \textbf{0.87} & 0.74 & \phantom{0}833.9 & \phantom{0}50.0 & 0.75 & 0.84 & \phantom{0}906.8 \\
     & \modelicon{chatgpt}GPT-6 Astra & \phantom{0}\textbf{90.0} & \underline{0.97} & \underline{0.90} & \textbf{1034.4} & \textbf{80.0} & \textbf{0.95} & \textbf{1.00} & \underline{1063.6} & \underline{70.0} & \underline{0.84} & \underline{0.97} & \textbf{1135.5} & \phantom{0}\textbf{90.0} & \underline{0.90} & 0.97 & \textbf{1103.0} \\
     & \modelicon{chatgpt}GPT-6.1 Sol & \phantom{0}\underline{80.0} & \textbf{1.00} & 0.78 & \phantom{0}991.5 & \textbf{80.0} & \textbf{0.95} & \underline{0.93} & \textbf{1111.2} & 60.0 & \underline{0.84} & 0.93 & \underline{1038.5} & \phantom{0}70.0 & 0.82 & \textbf{1.00} & \underline{1093.2} \\
    \midrule
    \multirow{4}{*}{\rotatebox[origin=c]{90}{\begin{tabular}{@{}c@{}}Text\\Only\end{tabular}}} & \modelicon{claude}Claude Opus 5.5 & \phantom{0}60.0 & \underline{0.77} & 0.80 & \phantom{0}889.2 & \textbf{40.0} & \textbf{0.93} & 0.45 & \underline{1000.0} & \phantom{0}\underline{0.0} & \textbf{0.60} & \textbf{0.93} & \underline{1000.0} & \phantom{0}20.0 & \underline{0.75} & 0.41 & \phantom{0}849.7 \\
     & \modelicon{claude}Claude Sonnet 5.5 & \phantom{0}\underline{80.0} & \textbf{1.00} & 0.71 & \phantom{0}908.3 & \underline{20.0} & 0.63 & 0.27 & \phantom{0}823.3 & \textbf{20.0} & \textbf{0.60} & 0.71 & \phantom{0}949.5 & \phantom{0}\textbf{60.0} & \textbf{0.95} & \underline{0.64} & \phantom{0}964.7 \\
     & \modelicon{chatgpt}GPT-6 Astra & \textbf{100.0} & \textbf{1.00} & \underline{0.91} & \textbf{1130.1} & \textbf{40.0} & \underline{0.80} & \textbf{0.90} & \textbf{1176.7} & \textbf{20.0} & \underline{0.50} & \underline{0.86} & \underline{1000.0} & \phantom{0}\underline{40.0} & 0.67 & \textbf{1.00} & \textbf{1130.7} \\
     & \modelicon{chatgpt}GPT-6.1 Sol & \textbf{100.0} & \textbf{1.00} & \textbf{1.00} & \underline{1072.4} & \underline{20.0} & 0.75 & \underline{0.73} & \underline{1000.0} & \textbf{20.0} & \textbf{0.60} & \underline{0.86} & \textbf{1050.5} & \phantom{0}20.0 & 0.55 & \textbf{1.00} & \underline{1054.9} \\
    \bottomrule
  \end{tabular}}
\end{table}

\subsection{Hand Error in Reconstruction}
\label{app:recon_hands}
Unlike generation, reconstruction is compared with ground-truth hands, and \Cref{tab:video} reports their error as MJE.
MJE keeps the global rotation and size of each hand, so it mixes where a hand is placed with how its fingers are posed.
We separate the two with Procrustes-aligned MJE (PA-MJE), which aligns each predicted hand to the GT by its best similarity transform in every frame before measuring the error~\cite{zimmermann2019freihand}.
\Cref{tab:app:recon_hands} shows that PA-MJE stays between 6.7 and 14.1 and varies much less across backbones than MJE.
This is clearest on co-dimensional clips, where MJE spreads from 17.6 to 33.4 but PA-MJE only from 9.9 to 14.1.
Most hand error thus lies in global rotation and scale rather than in finger pose.
\begin{table}[h]
\centering
\caption{Hand error in reconstruction by kind of matter, as the MJE of \Cref{tab:video} and PA-MJE. PA-MJE aligns each predicted hand to the GT by a similarity transform in every frame and thus removes the error in global rotation, translation and scale.}
\label{tab:app:recon_hands}\small
  \resizebox{0.85\linewidth}{!}{%
  \begin{tabular}{l|cc|cc|cc|cc}
    \toprule
    & \multicolumn{2}{c|}{\textcolor{matRigid}{\textbf{Rigid}}} & \multicolumn{2}{c|}{\textcolor{matDeform}{\textbf{Deformable}}} & \multicolumn{2}{c|}{\textcolor{matCodim}{\textbf{Co-dimensional}}} & \multicolumn{2}{c}{\textcolor{matFlow}{\textbf{Flowing}}} \\
    \cmidrule(lr){2-3}\cmidrule(lr){4-5}\cmidrule(lr){6-7}\cmidrule(lr){8-9}
    \multicolumn{1}{c|}{\textbf{Model}} & MJE ($\downarrow$) & PA-MJE ($\downarrow$) & MJE ($\downarrow$) & PA-MJE ($\downarrow$) & MJE ($\downarrow$) & PA-MJE ($\downarrow$) & MJE ($\downarrow$) & PA-MJE ($\downarrow$) \\
    \midrule
    \modelicon{claude}Claude Opus 5.5 & 29.1 & 13.1 & 26.0 & \underline{6.9} & 27.9 & 14.1 & 30.3 & 13.1 \\
    \modelicon{claude}Claude Sonnet 5.5 & 27.1 & 12.7 & 31.6 & 9.5 & \underline{27.5} & 11.1 & 28.3 & \underline{11.2} \\
    \modelicon{chatgpt}GPT-6 Astra & \textbf{26.0} & \textbf{11.1} & \underline{24.3} & 7.5 & 33.4 & \underline{10.7} & \textbf{21.7} & \textbf{11.1} \\
    \modelicon{chatgpt}GPT-6.1 Sol & \underline{26.4} & \underline{12.1} & \textbf{21.7} & \textbf{6.7} & \textbf{17.6} & \textbf{9.9} & \underline{24.2} & 12.5 \\
    \bottomrule
  \end{tabular}}
\end{table}

\FloatBarrier
\section{Limitations and Future Work}
\label{app:limits}
\noindent\textbf{Limitations.}\quad
While \method produces physically plausible interactions across diverse matter, several limitations warrant further investigation.
First, \method targets physical plausibility rather than physical accuracy.
Its materials start from presets and its hands are kinematic, so an interaction that looks plausible may still deviate from its real counterpart.
Second, \method simulates non-rigid matter with only four solvers (\ie, MPM, FEM, FLIP and XPBD), and distinct solvers act on each other in one direction only.
Matter and physical effects beyond the scope of these solvers therefore remain out of reach.
Third, transfer to another simulator still requires minor adjustments, such as attaching a grasped object to the robot hand because position control alone cannot regulate grip force.
Addressing these limitations would bring \method closer to physically accurate interactions that carry over to any simulator and any kind of matter without even minor adjustments.

\vspace{0.5em}\noindent\textbf{Future work.}\quad
\method opens up several promising avenues for future research.
First, since the agents call perception models as tools, stronger or commercial models and retrieved assets can be plugged in much as in retrieval-augmented generation to yield more detailed and plausible interactions.
Second, training policies on the transferred rollouts with reinforcement learning~\cite{rajeswaran2018dapg} or vision-language-action models~\cite{kim2024openvla} could supply the grip regulation that position control lacks and extend dexterous manipulation to non-rigid matter.
Third, adding solvers beyond the current four and coupling them in both directions would extend \method to a wider range of matter and interactions.
Progress in these directions would bring \method closer to plausible and transferable interactions with any kind of matter.

\raggedbottom
\ifdefined\nolinenumbers\nolinenumbers\fi
\section{Benchmark Catalogue}
\label{app:catalogue}
In this section, we catalogue the \benchTasks{} generation tasks of \bench in full so that each task and its scoring can be inspected and reproduced.
We first list each task with its instruction, input and hidden annotations in \Cref{app:cat:tasks}.
We then define the metric behind each kind of success condition in \Cref{app:cat:metrics}.
Finally, we list the physical checks shared by all tasks in \Cref{app:cat:checks}.
The agents and the perceptual judge do not see the success conditions and part roles, whereas the checker also applies the physical checks during a run to guide the agents.

\definecolor{krigid}{HTML}{2A78D6}\definecolor{kdeformable}{HTML}{EB6834}
\definecolor{kcodimensional}{HTML}{1BAF7A}\definecolor{kflowing}{HTML}{EDA100}
\definecolor{ktext}{HTML}{7C90E0}
\newcommand{\textonlypanel}{\begin{tikzpicture}[scale=0.767]
  \useasboundingbox (0,0) rectangle (1.76,-1.32);
  \draw[black!78, line width=0.35pt, fill=white] (0,0) rectangle (1.76,-1.32);
  \draw[ktext, line width=0.6pt, fill=ktext!12, rounded corners=1.5pt]
    (0.45,-0.41) -- (1.31,-0.41) -- (1.31,-0.91) -- (0.71,-0.91) -- (0.55,-1.06) -- (0.57,-0.91) -- (0.45,-0.91) -- cycle;
  \foreach \y/\l in {-0.55/0.62, -0.66/0.62, -0.77/0.4} \draw[ktext, line width=0.5pt, line cap=round] (0.57,\y) -- ++(\l,0);
\end{tikzpicture}}

\subsection{Tasks and Hidden Annotations}
\label{app:cat:tasks}
Each generation task pairs an instruction for the agents with annotations that only the evaluation sees.
\Cref{tab:cat_tasks} lists every instruction verbatim with its input and these hidden annotations, grouped by kind of matter.
A task ID combines the input (S for a 3D scene, T for text alone and I for an image) with the kind of matter (R, D, C and F).
The hidden annotations consist of success conditions and part roles.
Success conditions compare the initial and final states of the named objects unless they measure the motion itself.
Part roles name the parts that the hands must touch, must keep clear of or should preferably avoid.

{\scriptsize
\setlength{\tabcolsep}{3pt}\renewcommand{\arraystretch}{1.2}
\begin{longtable}{@{}>{\raggedright\arraybackslash}p{\dimexpr 0.08\linewidth-2\tabcolsep\relax}>{\raggedright\arraybackslash}p{\dimexpr 0.05\linewidth-2\tabcolsep\relax}>{\raggedright\arraybackslash}p{\dimexpr 0.33\linewidth-2\tabcolsep\relax}>{\raggedright\arraybackslash}p{\dimexpr 0.54\linewidth-2\tabcolsep\relax}@{}}
  \caption{\small Generation tasks of \bench by kind of matter, each with its input, instruction and hidden annotations. The agents see only the input and the instruction, whereas the success conditions and the part roles below them score the result.}
  \label{tab:cat_tasks}\\
  \toprule
  \textbf{Input} & \textbf{Task} & \textbf{Instruction} & \textbf{Hidden Annotations} \\
  \midrule
  \endfirsthead
  \multicolumn{4}{@{}l}{\small\emph{\Cref{tab:cat_tasks}, continued}} \\
  \toprule
  \textbf{Input} & \textbf{Task} & \textbf{Instruction} & \textbf{Hidden Annotations} \\
  \midrule
  \endhead
  \bottomrule
  \endfoot
  \rowcolor{krigid!14}\multicolumn{4}{@{\hspace{2pt}}l}{\rule{0pt}{9pt}\textcolor{krigid}{\rule{5pt}{5pt}}\hspace{4pt}\textbf{Rigid}} \\
\vtop{\vskip-1.9ex\hbox{\includegraphics[width=1.35cm]{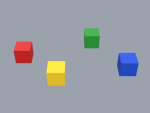}}} & \textbf{SR1} & ``Stack the red, green and blue cubes with red at the bottom and blue on top; leave the yellow cube where it is.'' & \textbullet\,the objects stand stacked, bottom to top: \texttt{red\_\allowbreak{}cube} $\rightarrow$ \texttt{green\_\allowbreak{}cube} $\rightarrow$ \texttt{blue\_\allowbreak{}cube} \newline \textbullet\,\texttt{blue\_\allowbreak{}cube} is let go at the end \newline \textbullet\,\texttt{blue\_\allowbreak{}cube} is at rest at the end ($\leq$\,5\,cm/s) \newline \textbullet\,\texttt{yellow\_\allowbreak{}cube} moves $\leq$\,1\,cm \\
\rowcolor{black!3}\vtop{\vskip-1.9ex\hbox{\includegraphics[width=1.35cm]{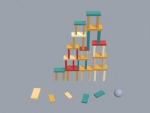}}} & \textbf{SR2} & ``Take the cap block off the top of the tower without knocking any other block off.'' & \textbullet\,\texttt{block\_\allowbreak{}41} moves on the table $\geq$\,6\,cm \newline \textbullet\,\texttt{block\_\allowbreak{}39} moves on the table $\leq$\,0.5\,cm \newline \textbullet\,\texttt{block\_\allowbreak{}40} moves on the table $\leq$\,0.5\,cm \newline \textbullet\,\texttt{block\_\allowbreak{}00} moves on the table $\leq$\,0.5\,cm \\
\vtop{\vskip-1.9ex\hbox{\includegraphics[width=1.35cm]{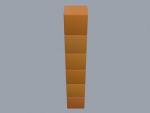}}} & \textbf{SR3} & ``Carry the tower of cubes 15 cm to the right without it falling apart.'' & \textbullet\,\texttt{cube\_\allowbreak{}1} moves to the right $\geq$\,12\,cm \newline \textbullet\,\texttt{cube\_\allowbreak{}6} moves to the right $\geq$\,12\,cm \newline \textbullet\,the objects stand stacked, bottom to top: \texttt{cube\_\allowbreak{}1} $\rightarrow$ \texttt{cube\_\allowbreak{}2} $\rightarrow$ \texttt{cube\_\allowbreak{}3} $\rightarrow$ \texttt{cube\_\allowbreak{}4} $\rightarrow$ \texttt{cube\_\allowbreak{}5} $\rightarrow$ \texttt{cube\_\allowbreak{}6} \\
\rowcolor{black!3}\vtop{\vskip-1.9ex\hbox{\includegraphics[width=1.35cm]{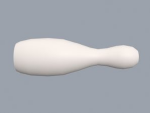}}} & \textbf{SR4} & ``Stand the fallen bowling pin back up.'' & \textbullet\,\texttt{pin} ends pointing up, off by $\leq$\,10$^\circ$ \newline \textbullet\,\texttt{pin} is let go at the end \newline \textbullet\,\texttt{pin} is at rest at the end ($\leq$\,5\,cm/s) \\
\vtop{\vskip-1.9ex\hbox{\includegraphics[width=1.35cm]{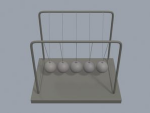}}} & \textbf{SR5} & ``Pull the ball at the left end out to the side and let it go, so that the ball at the right end swings out.'' & \textbullet\,\texttt{ball\_\allowbreak{}1} rises $\geq$\,2\,cm \newline \textbullet\,\texttt{ball\_\allowbreak{}1} is let go at the end \newline \textbullet\,\texttt{ball\_\allowbreak{}5} rises $\geq$\,1\,cm \newline \textbullet\,hands stay 0.3\,cm clear of \texttt{ball\_\allowbreak{}5} \\
\rowcolor{black!3}\vtop{\vskip-1.9ex\hbox{\includegraphics[width=1.35cm]{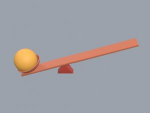}}} & \textbf{SR6} & ``Push down the high end of the seesaw so that the ball flies up.'' & \textbullet\,\texttt{ball} rises $\geq$\,5\,cm touching nothing \newline \textbullet\,\texttt{seesaw.\allowbreak{}high\_\allowbreak{}end} is held by the hand \newline \textbullet\,hands touch \texttt{seesaw.\allowbreak{}high\_\allowbreak{}end} \newline \textbullet\,hands stay 0.3\,cm clear of \texttt{ball} \\
\vtop{\vskip-1.9ex\hbox{\includegraphics[width=1.35cm]{fig/bench/arch_tower.pdf}}} & \textbf{SR7} & ``Knock the tower down by throwing the ball at it.'' & \textbullet\,the lowest point of \texttt{block\_\allowbreak{}41} ends $\leq$\,5\,cm up \newline \textbullet\,\texttt{ball} hits the \texttt{block} after leaving the hand \newline \textbullet\,\texttt{ball} travels $\geq$\,10\,cm with no hand on it \\
\rowcolor{black!3}\vtop{\vskip-1.9ex\hbox{\includegraphics[width=1.35cm]{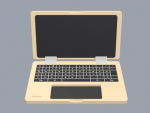}}} & \textbf{SR8} & ``Pack the laptop away: close it and pick it up.'' & \textbullet\,\texttt{laptop.\allowbreak{}lid} turns $\geq$\,95$^\circ$ on \texttt{laptop.\allowbreak{}base} (at the end) \newline \textbullet\,\texttt{laptop} rises $\geq$\,5\,cm \\
\vtop{\vskip-1.9ex\hbox{\includegraphics[width=1.35cm]{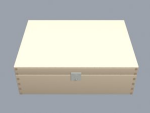}}} & \textbf{SR9} & ``Flip the latch open, then hold the box with your left hand and lift its lid with your right.'' & \textbullet\,\texttt{box.\allowbreak{}latch\_\allowbreak{}hasp} turns $\geq$\,60$^\circ$ on \texttt{box.\allowbreak{}lid} (at most during the motion) \newline \textbullet\,\texttt{box.\allowbreak{}lid} turns $\geq$\,60$^\circ$ on \texttt{box.\allowbreak{}base} (at the end) \newline \textbullet\,\texttt{box.\allowbreak{}base} is held by the left hand \newline \textbullet\,\texttt{box.\allowbreak{}lid} is held by the right hand \newline \textbullet\,both hands touch the targets \\
\rowcolor{black!3}\vtop{\vskip-1.9ex\hbox{\includegraphics[width=1.35cm]{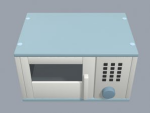}}} & \textbf{SR10} & ``Open the microwave door, then turn the timer knob a quarter turn.'' & \textbullet\,\texttt{microwave.\allowbreak{}door} turns $\geq$\,60$^\circ$ on \texttt{microwave.\allowbreak{}body} (at the end) \newline \textbullet\,\texttt{microwave.\allowbreak{}knob} turns $\geq$\,80$^\circ$ on \texttt{microwave.\allowbreak{}body} (at the end) \\
\vtop{\vskip-1.9ex\hbox{\textonlypanel}} & \textbf{TR1} & ``Tip the first domino so that the whole row falls.'' & \textbullet\,\texttt{domino\_\allowbreak{}1} ends pointing up, off by $\geq$\,30$^\circ$ \newline \textbullet\,\texttt{domino\_\allowbreak{}2} ends pointing up, off by $\geq$\,30$^\circ$ \newline \textbullet\,\texttt{domino\_\allowbreak{}3} ends pointing up, off by $\geq$\,30$^\circ$ \newline \textbullet\,\texttt{domino\_\allowbreak{}4} ends pointing up, off by $\geq$\,30$^\circ$ \newline \textbullet\,\texttt{domino\_\allowbreak{}5} ends pointing up, off by $\geq$\,30$^\circ$ \newline \textbullet\,\texttt{domino\_\allowbreak{}6} ends pointing up, off by $\geq$\,30$^\circ$ \newline \textbullet\,hands stay 0.3\,cm clear of \texttt{domino\_\allowbreak{}2} \newline \textbullet\,hands stay 0.3\,cm clear of \texttt{domino\_\allowbreak{}3} \newline \textbullet\,hands stay 0.3\,cm clear of \texttt{domino\_\allowbreak{}4} \newline \textbullet\,hands stay 0.3\,cm clear of \texttt{domino\_\allowbreak{}5} \newline \textbullet\,hands stay 0.3\,cm clear of \texttt{domino\_\allowbreak{}6} \\
\rowcolor{black!3}\vtop{\vskip-1.9ex\hbox{\textonlypanel}} & \textbf{TR2} & ``Stack the books into one pile, the largest at the bottom and the smallest on top.'' & \textbullet\,the objects stand stacked, bottom to top: \texttt{book\_\allowbreak{}large} $\rightarrow$ \texttt{book\_\allowbreak{}medium} $\rightarrow$ \texttt{book\_\allowbreak{}small} $\rightarrow$ \texttt{book\_\allowbreak{}tiny} \newline \textbullet\,\texttt{book\_\allowbreak{}tiny} is let go at the end \newline \textbullet\,\texttt{book\_\allowbreak{}tiny} is at rest at the end ($\leq$\,5\,cm/s) \\
\vtop{\vskip-1.9ex\hbox{\textonlypanel}} & \textbf{TR3} & ``Flip the cutting board over with both hands.'' & \textbullet\,\texttt{board} ends pointing [0, 0, -1], off by $\leq$\,20$^\circ$ \newline \textbullet\,\texttt{board} is let go at the end \newline \textbullet\,\texttt{board} is at rest at the end ($\leq$\,5\,cm/s) \newline \textbullet\,both hands touch the targets \\
\rowcolor{black!3}\vtop{\vskip-1.9ex\hbox{\textonlypanel}} & \textbf{TR4} & ``Use the opener to pop the cap off the bottle.'' & \textbullet\,\texttt{cap} ends $\geq$\,1\,cm away from the \texttt{bottle} \newline \textbullet\,\texttt{opener} is held by the hand \newline \textbullet\,hands stay 0.3\,cm clear of \texttt{cap} \\
\vtop{\vskip-1.9ex\hbox{\textonlypanel}} & \textbf{TR5} & ``Open the toolbox, take the hammer out by its handle and lay it on the table to the right of the box.'' & \textbullet\,\texttt{toolbox.\allowbreak{}lid} turns $\geq$\,60$^\circ$ on \texttt{toolbox.\allowbreak{}base} (at most during the motion) \newline \textbullet\,$\leq$\,10\% of \texttt{hammer} ends in or on the \texttt{toolbox} \newline \textbullet\,\texttt{hammer} moves to the right $\geq$\,10\,cm \newline \textbullet\,\texttt{hammer} is let go at the end \newline \textbullet\,\texttt{hammer} is at rest at the end ($\leq$\,5\,cm/s) \newline \textbullet\,hands touch \texttt{hammer.\allowbreak{}handle} \\
\rowcolor{black!3}\vtop{\vskip-1.9ex\hbox{\includegraphics[width=1.35cm]{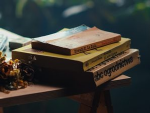}}} & \textbf{IR1} & ``Pull the yellow book out from the middle of the stack without the top book falling off.'' & \textbullet\,\texttt{book\_\allowbreak{}middle} moves on the table $\geq$\,12\,cm \newline \textbullet\,the objects stand stacked, bottom to top: \texttt{book\_\allowbreak{}bottom} $\rightarrow$ \texttt{book\_\allowbreak{}top} \newline \textbullet\,\texttt{book\_\allowbreak{}bottom} moves on the table $\leq$\,2\,cm \\
\vtop{\vskip-1.9ex\hbox{\includegraphics[width=1.35cm]{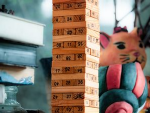}}} & \textbf{IR2} & ``Push the block marked 11 out of the tower without the tower falling.'' & \textbullet\,\texttt{block\_\allowbreak{}11} moves on the table $\geq$\,5\,cm \newline \textbullet\,\texttt{tower} moves on the table $\leq$\,0.5\,cm \newline \textbullet\,\texttt{tower} tilts $\leq$\,3$^\circ$ \newline \textbullet\,hands stay 0.3\,cm clear of \texttt{tower} \\
\rowcolor{black!3}\vtop{\vskip-1.9ex\hbox{\includegraphics[width=1.35cm]{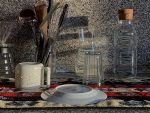}}} & \textbf{IR3} & ``Turn the saucer right side up and set the mug on it.'' & \textbullet\,\texttt{saucer} ends pointing up, off by $\geq$\,150$^\circ$ \newline \textbullet\,$\geq$\,90\% of \texttt{mug} ends in or on the \texttt{saucer} \newline \textbullet\,\texttt{mug} is let go at the end \newline \textbullet\,\texttt{mug} is at rest at the end ($\leq$\,5\,cm/s) \newline \textbullet\,\texttt{saucer} is let go at the end \newline \textbullet\,\texttt{glass} moves on the table $\leq$\,1\,cm \newline \textbullet\,\texttt{carafe} moves on the table $\leq$\,1\,cm \\
\vtop{\vskip-1.9ex\hbox{\includegraphics[width=1.35cm]{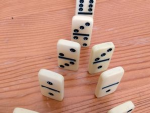}}} & \textbf{IR4} & ``Tip the domino at the far end into the next one so that both lines fall.'' & \textbullet\,\texttt{domino\_\allowbreak{}1} ends pointing up, off by $\geq$\,12$^\circ$ \newline \textbullet\,\texttt{domino\_\allowbreak{}2} ends pointing up, off by $\geq$\,12$^\circ$ \newline \textbullet\,\texttt{domino\_\allowbreak{}3} ends pointing up, off by $\geq$\,12$^\circ$ \newline \textbullet\,\texttt{domino\_\allowbreak{}4} ends pointing up, off by $\geq$\,12$^\circ$ \newline \textbullet\,\texttt{domino\_\allowbreak{}5} ends pointing up, off by $\geq$\,12$^\circ$ \newline \textbullet\,\texttt{domino\_\allowbreak{}6} ends pointing up, off by $\geq$\,12$^\circ$ \newline \textbullet\,\texttt{domino\_\allowbreak{}7} ends pointing up, off by $\geq$\,12$^\circ$ \newline \textbullet\,\texttt{domino\_\allowbreak{}8} ends pointing up, off by $\geq$\,12$^\circ$ \newline \textbullet\,\texttt{domino\_\allowbreak{}9} ends pointing up, off by $\geq$\,12$^\circ$ \newline \textbullet\,hands stay 0.3\,cm clear of \texttt{domino\_\allowbreak{}2} \newline \textbullet\,hands stay 0.3\,cm clear of \texttt{domino\_\allowbreak{}3} \newline \textbullet\,hands stay 0.3\,cm clear of \texttt{domino\_\allowbreak{}4} \newline \textbullet\,hands stay 0.3\,cm clear of \texttt{domino\_\allowbreak{}5} \newline \textbullet\,hands stay 0.3\,cm clear of \texttt{domino\_\allowbreak{}6} \newline \textbullet\,hands stay 0.3\,cm clear of \texttt{domino\_\allowbreak{}7} \newline \textbullet\,hands stay 0.3\,cm clear of \texttt{domino\_\allowbreak{}8} \newline \textbullet\,hands stay 0.3\,cm clear of \texttt{domino\_\allowbreak{}9} \\
\rowcolor{black!3}\vtop{\vskip-1.9ex\hbox{\includegraphics[width=1.35cm]{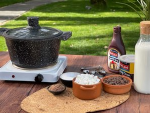}}} & \textbf{IR5} & ``Lift the lid off the pot by its knob and set it down on the table.'' & \textbullet\,\texttt{lid} ends $\geq$\,2\,cm away from the \texttt{pot} \newline \textbullet\,the lowest point of \texttt{lid} ends $\leq$\,1\,cm up \newline \textbullet\,\texttt{lid} is let go at the end \newline \textbullet\,\texttt{lid} is at rest at the end ($\leq$\,5\,cm/s) \newline \textbullet\,\texttt{lid.\allowbreak{}knob} is held by the hand \newline \textbullet\,\texttt{pot} moves on the table $\leq$\,1\,cm \newline \textbullet\,\texttt{marshmallow\_\allowbreak{}bowl} moves on the table $\leq$\,1\,cm \newline \textbullet\,hands touch \texttt{lid.\allowbreak{}knob} \\
\rowcolor{kdeformable!14}\multicolumn{4}{@{\hspace{2pt}}l}{\rule{0pt}{9pt}\textcolor{kdeformable}{\rule{5pt}{5pt}}\hspace{4pt}\textbf{Deformable}} \\
\vtop{\vskip-1.9ex\hbox{\includegraphics[width=1.35cm]{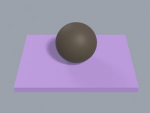}}} & \textbf{SD1} & ``Press the dough with your fingertips, then let go.'' & \textbullet\,\texttt{dough} keeps a dent $\geq$\,0.8\,cm deep \newline \textbullet\,\texttt{dough} is let go at the end \\
\rowcolor{black!3}\vtop{\vskip-1.9ex\hbox{\includegraphics[width=1.35cm]{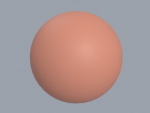}}} & \textbf{SD2} & ``Press the ball with your fingertips, then let go.'' & \textbullet\,\texttt{soft\_\allowbreak{}ball} is dented $\geq$\,0.8\,cm deep \newline \textbullet\,\texttt{soft\_\allowbreak{}ball} springs back to $\geq$\,0.95 of its width \newline \textbullet\,\texttt{soft\_\allowbreak{}ball} is let go at the end \\
\vtop{\vskip-1.9ex\hbox{\includegraphics[width=1.35cm]{fig/bench/dough.pdf}}} & \textbf{SD3} & ``Make a hollow in the middle of the dough so that it can hold an egg.'' & \textbullet\,\texttt{dough} keeps a dent $\geq$\,1.5\,cm deep \newline \textbullet\,\texttt{dough} holds one hollow of $\geq$\,6\,cm$^2$ \newline \textbullet\,\texttt{dough} ends at $\geq$\,0.7 of its height \newline \textbullet\,\texttt{dough} is let go at the end \\
\rowcolor{black!3}\vtop{\vskip-1.9ex\hbox{\includegraphics[width=1.35cm]{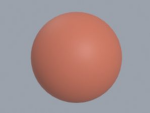}}} & \textbf{SD4} & ``Roll the clay between your palms into a sausage.'' & \textbullet\,\texttt{clay} ends $\geq$\,2 times as long as wide \newline \textbullet\,\texttt{clay} changes its volume by $\leq$\,15\% \newline \textbullet\,both hands touch the targets \\
\vtop{\vskip-1.9ex\hbox{\includegraphics[width=1.35cm]{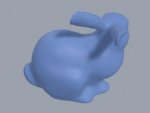}}} & \textbf{SD5} & ``Gently pick up the bunny and make it hop along the table, like a real bunny.'' & \textbullet\,\texttt{bunny} hops $\geq$\,2 times \newline \textbullet\,\texttt{bunny} moves on the table $\geq$\,10\,cm \newline \textbullet\,\texttt{bunny} is pressed $\leq$\,20\% thinner \\
\rowcolor{black!3}\vtop{\vskip-1.9ex\hbox{\includegraphics[width=1.35cm]{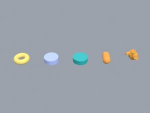}}} & \textbf{SD6} & ``Squeeze the soft ring between your thumb and index finger until its hole closes, then let it go.'' & \textbullet\,\texttt{ring} is pressed $\geq$\,35\% thinner \newline \textbullet\,\texttt{ring} is held by the thumb and index together \newline \textbullet\,\texttt{ring} springs back to $\geq$\,0.9 of its width \newline \textbullet\,\texttt{ring} is let go at the end \\
\vtop{\vskip-1.9ex\hbox{\includegraphics[width=1.35cm]{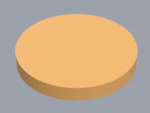}}} & \textbf{SD7} & ``Break the cookie in half.'' & \textbullet\,\texttt{cookie} ends in 2 pieces (each 30\% of it or more) \\
\rowcolor{black!3}\vtop{\vskip-1.9ex\hbox{\includegraphics[width=1.35cm]{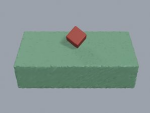}}} & \textbf{SD8} & ``Bury the red cube halfway in the goo.'' & \textbullet\,the lowest point of \texttt{cube} ends $\leq$\,3.5\,cm up \newline \textbullet\,\texttt{cube} is let go at the end \\
\vtop{\vskip-1.9ex\hbox{\includegraphics[width=1.35cm]{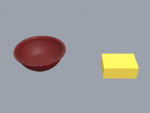}}} & \textbf{SD9} & ``Squeeze the water out of the sponge into the bowl, then put the sponge down next to the bowl.'' & \textbullet\,$\geq$\,30\% of \texttt{water} ends inside the \texttt{bowl} \newline \textbullet\,\texttt{sponge} is pressed $\geq$\,30\% thinner \newline \textbullet\,$\leq$\,20\% of \texttt{sponge} ends in or on the \texttt{bowl} \newline \textbullet\,\texttt{sponge} is let go at the end \\
\rowcolor{black!3}\vtop{\vskip-1.9ex\hbox{\includegraphics[width=1.35cm]{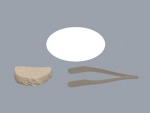}}} & \textbf{SD10} & ``Use the tongs to put the bread on the plate.'' & \textbullet\,$\geq$\,80\% of \texttt{bread} ends in or on the \texttt{plate} \newline \textbullet\,\texttt{tongs} is held by the hand \newline \textbullet\,hands stay 0.5\,cm clear of \texttt{bread} \\
\vtop{\vskip-1.9ex\hbox{\textonlypanel}} & \textbf{TD1} & ``Make a snowman: set the small ball on top of the big one and press it down a little so that it sticks.'' & \textbullet\,the objects stand stacked, bottom to top: \texttt{big\_\allowbreak{}ball} $\rightarrow$ \texttt{small\_\allowbreak{}ball} \newline \textbullet\,\texttt{big\_\allowbreak{}ball} ends at $\geq$\,0.8 of its height \newline \textbullet\,\texttt{small\_\allowbreak{}ball} is let go at the end \newline \textbullet\,\texttt{small\_\allowbreak{}ball} is at rest at the end ($\leq$\,5\,cm/s) \\
\rowcolor{black!3}\vtop{\vskip-1.9ex\hbox{\textonlypanel}} & \textbf{TD2} & ``Put the teddy bear away in the jar.'' & \textbullet\,$\geq$\,80\% of \texttt{bear} ends inside the \texttt{jar} \newline \textbullet\,\texttt{bear} is let go at the end \\
\vtop{\vskip-1.9ex\hbox{\textonlypanel}} & \textbf{TD3} & ``Pull the slime between your hands until it is twice as long as it was.'' & \textbullet\,\texttt{slime} stretches to $\geq$\,2 times its length \newline \textbullet\,both hands touch the targets \\
\rowcolor{black!3}\vtop{\vskip-1.9ex\hbox{\textonlypanel}} & \textbf{TD4} & ``Squeeze some ketchup onto the middle of the plate.'' & \textbullet\,$\geq$\,10\% of \texttt{ketchup} ends in or on the \texttt{plate} \newline \textbullet\,\texttt{bottle} is pressed $\geq$\,15\% thinner \newline \textbullet\,$\leq$\,10\% of \texttt{ketchup} spills outside the \texttt{bottle} and \texttt{plate} \\
\vtop{\vskip-1.9ex\hbox{\textonlypanel}} & \textbf{TD5} & ``Scoop a ball of ice cream into the bowl.'' & \textbullet\,\texttt{ice\_\allowbreak{}cream} ends in $\geq$\,2 pieces (each 3\% of it or more) \newline \textbullet\,$\geq$\,3\% of \texttt{ice\_\allowbreak{}cream} ends in or on the \texttt{bowl} \newline \textbullet\,\texttt{scoop.\allowbreak{}handle} is held by the hand \newline \textbullet\,hands stay 0.5\,cm clear of \texttt{ice\_\allowbreak{}cream} \\
\rowcolor{black!3}\vtop{\vskip-1.9ex\hbox{\includegraphics[width=1.35cm]{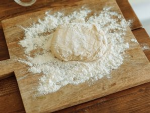}}} & \textbf{ID1} & ``Press a deep well into the middle of the dough with your fingers.'' & \textbullet\,\texttt{dough} keeps a dent $\geq$\,1.5\,cm deep \newline \textbullet\,\texttt{dough} ends at $\geq$\,0.7 of its height \newline \textbullet\,\texttt{dough} is let go at the end \\
\vtop{\vskip-1.9ex\hbox{\includegraphics[width=1.35cm]{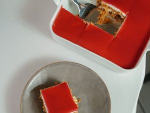}}} & \textbf{ID2} & ``Lift the square nearest the plate out of the dish with the server and put it on the plate beside the other one.'' & \textbullet\,$\geq$\,90\% of \texttt{square} ends in or on the \texttt{plate} \newline \textbullet\,\texttt{square} stays in one piece (pieces of 20\% of it or more counted) \newline \textbullet\,\texttt{server.\allowbreak{}handle} is held by the hand \newline \textbullet\,\texttt{slice} moves on the table $\leq$\,1\,cm \newline \textbullet\,\texttt{served} moves on the table $\leq$\,2\,cm \newline \textbullet\,hands stay 0.3\,cm clear of \texttt{square} \newline \textbullet\,hands touch \texttt{server.\allowbreak{}handle} \\
\rowcolor{black!3}\vtop{\vskip-1.9ex\hbox{\includegraphics[width=1.35cm]{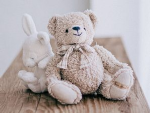}}} & \textbf{ID3} & ``Gently lift the bunny and sit it on the teddy bear's lap.'' & \textbullet\,$\geq$\,60\% of \texttt{bunny} ends in or on the \texttt{teddy} \newline \textbullet\,\texttt{bunny} is pressed $\leq$\,25\% thinner \newline \textbullet\,\texttt{bunny} is let go at the end \newline \textbullet\,\texttt{bunny} is at rest at the end ($\leq$\,5\,cm/s) \newline \textbullet\,\texttt{teddy} moves on the table $\leq$\,3\,cm \newline \textbullet\,\texttt{teddy} ends pointing up, off by $\leq$\,20$^\circ$ \\
\vtop{\vskip-1.9ex\hbox{\includegraphics[width=1.35cm]{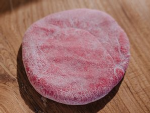}}} & \textbf{ID4} & ``Roll the play dough on the table into a long, thin snake.'' & \textbullet\,\texttt{play\_\allowbreak{}dough} ends $\geq$\,1.8 times its length \newline \textbullet\,\texttt{play\_\allowbreak{}dough} ends $\geq$\,6 times as long as wide \newline \textbullet\,\texttt{play\_\allowbreak{}dough} changes its volume by $\leq$\,15\% \\
\rowcolor{black!3}\vtop{\vskip-1.9ex\hbox{\includegraphics[width=1.35cm]{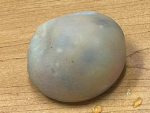}}} & \textbf{ID5} & ``Squeeze the grey stress ball hard, then let it go.'' & \textbullet\,\texttt{stress\_\allowbreak{}ball} is pressed $\geq$\,30\% thinner \newline \textbullet\,\texttt{stress\_\allowbreak{}ball} springs back to $\geq$\,0.9 of its width \newline \textbullet\,\texttt{stress\_\allowbreak{}ball} is let go at the end \\
\rowcolor{kcodimensional!14}\multicolumn{4}{@{\hspace{2pt}}l}{\rule{0pt}{9pt}\textcolor{kcodimensional}{\rule{5pt}{5pt}}\hspace{4pt}\textbf{Co-dimensional}} \\
\vtop{\vskip-1.9ex\hbox{\includegraphics[width=1.35cm]{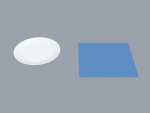}}} & \textbf{SC1} & ``Fold the towel in half and put it on the plate.'' & \textbullet\,\texttt{towel} ends on $\leq$\,0.65 of its first footprint \newline \textbullet\,$\geq$\,80\% of \texttt{towel} ends in or on the \texttt{plate} \\
\rowcolor{black!3}\vtop{\vskip-1.9ex\hbox{\includegraphics[width=1.35cm]{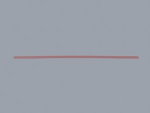}}} & \textbf{SC2} & ``Fold the ribbon back and forth into a small zigzag stack.'' & \textbullet\,\texttt{ribbon} ends on $\leq$\,0.4 of its first footprint \newline \textbullet\,\texttt{ribbon} gathers within a radius of $\leq$\,6\,cm (90\%) \newline \textbullet\,$\geq$\,60\% of \texttt{ribbon} lies flat along one line \\
\vtop{\vskip-1.9ex\hbox{\includegraphics[width=1.35cm]{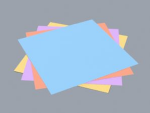}}} & \textbf{SC3} & ``Take only the top cloth off the stack and lay it on the table to the right.'' & \textbullet\,\texttt{cloth\_\allowbreak{}5} moves to the right $\geq$\,15\,cm \newline \textbullet\,\texttt{cloth\_\allowbreak{}4} moves on the table $\leq$\,3\,cm \\
\rowcolor{black!3}\vtop{\vskip-1.9ex\hbox{\includegraphics[width=1.35cm]{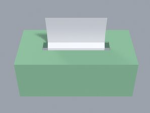}}} & \textbf{SC4} & ``Pull one tissue out of the box for me.'' & \textbullet\,only one \texttt{tissue\_\allowbreak{}*} ends $\geq$\,5 cm away from the \texttt{box} \newline \textbullet\,\texttt{tissue\_\allowbreak{}1} ends $\geq$\,5\,cm away from the \texttt{box} \newline \textbullet\,hands touch \texttt{tissue\_\allowbreak{}1} \\
\vtop{\vskip-1.9ex\hbox{\includegraphics[width=1.35cm]{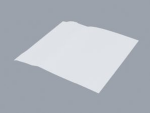}}} & \textbf{SC5} & ``Fold the paper in half and press along the fold.'' & \textbullet\,\texttt{paper} has a sharp fold \newline \textbullet\,\texttt{paper} ends on 0.4--0.65 of its first footprint \newline \textbullet\,$\leq$\,5\% of \texttt{paper} wrinkles \\
\rowcolor{black!3}\vtop{\vskip-1.9ex\hbox{\includegraphics[width=1.35cm]{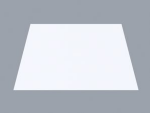}}} & \textbf{SC6} & ``Crumple the foil into a ball and put it down.'' & \textbullet\,\texttt{foil} ends on $\leq$\,0.1 of its first footprint \newline \textbullet\,$\geq$\,30\% of \texttt{foil} crumples \newline \textbullet\,\texttt{foil} is let go at the end \newline \textbullet\,\texttt{foil} is at rest at the end ($\leq$\,5\,cm/s) \\
\vtop{\vskip-1.9ex\hbox{\includegraphics[width=1.35cm]{fig/bench/paper.pdf}}} & \textbf{SC7} & ``Tear the paper into two pieces.'' & \textbullet\,\texttt{paper} ends in 2 pieces (each 20\% of it or more) \\
\rowcolor{black!3}\vtop{\vskip-1.9ex\hbox{\includegraphics[width=1.35cm]{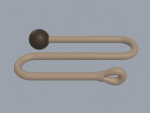}}} & \textbf{SC8} & ``Lift the rope until the weight hangs off the table.'' & \textbullet\,the lowest point of \texttt{weight} ends $\geq$\,5\,cm up \newline \textbullet\,hands touch \texttt{rope} \newline \textbullet\,hands preferably stay off \texttt{weight} \\
\vtop{\vskip-1.9ex\hbox{\includegraphics[width=1.35cm]{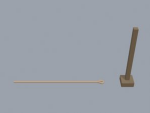}}} & \textbf{SC9} & ``Wind the rope three times around the post.'' & \textbullet\,\texttt{rope} wraps $\geq$\,1000$^\circ$ around the \texttt{post} \\
\rowcolor{black!3}\vtop{\vskip-1.9ex\hbox{\includegraphics[width=1.35cm]{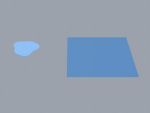}}} & \textbf{SC10} & ``Wipe up the spill with the towel.'' & \textbullet\,\texttt{towel} wipes over $\geq$\,80\% of the \texttt{spill} \\
\vtop{\vskip-1.9ex\hbox{\textonlypanel}} & \textbf{TC1} & ``Fold the T-shirt into a neat rectangle: sleeves in, then in half.'' & \textbullet\,\texttt{tshirt} ends on $\leq$\,0.4 of its first footprint \newline \textbullet\,\texttt{tshirt} is let go at the end \\
\rowcolor{black!3}\vtop{\vskip-1.9ex\hbox{\textonlypanel}} & \textbf{TC2} & ``Tie the scarf once around the neck of the bottle.'' & \textbullet\,\texttt{scarf} wraps $\geq$\,330$^\circ$ around the \texttt{bottle} \newline \textbullet\,the lowest point of \texttt{scarf} ends $\geq$\,5\,cm up \\
\vtop{\vskip-1.9ex\hbox{\textonlypanel}} & \textbf{TC3} & ``Wrap the box in the paper.'' & \textbullet\,\texttt{paper} covers $\geq$\,90\% of the top and sides of the \texttt{box} \newline \textbullet\,\texttt{paper} is let go at the end \\
\rowcolor{black!3}\vtop{\vskip-1.9ex\hbox{\textonlypanel}} & \textbf{TC4} & ``Coil the cable into a neat loop and leave it on the table.'' & \textbullet\,\texttt{cable} gathers within a radius of $\leq$\,12\,cm (90\%) \newline \textbullet\,\texttt{cable} is let go at the end \\
\vtop{\vskip-1.9ex\hbox{\textonlypanel}} & \textbf{TC5} & ``Crumple the foil into a ball and throw it into the bin.'' & \textbullet\,$\geq$\,90\% of \texttt{foil} ends inside the \texttt{bin} \newline \textbullet\,\texttt{foil} ends on $\leq$\,0.1 of its first footprint \newline \textbullet\,\texttt{foil} travels $\geq$\,15\,cm with no hand on it \\
\rowcolor{black!3}\vtop{\vskip-1.9ex\hbox{\includegraphics[width=1.35cm]{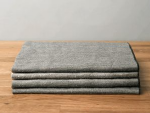}}} & \textbf{IC1} & ``Take the top towel, unfold it and spread it flat on the table.'' & \textbullet\,\texttt{top\_\allowbreak{}towel} ends on $\geq$\,2.5 of its first footprint \newline \textbullet\,$\leq$\,10\% of \texttt{top\_\allowbreak{}towel} is lifted off the table \newline \textbullet\,\texttt{stack} moves on the table $\leq$\,2\,cm \\
\vtop{\vskip-1.9ex\hbox{\includegraphics[width=1.35cm]{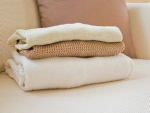}}} & \textbf{IC2} & ``Pull the brown sweater out from the middle without the cream one on top falling off.'' & \textbullet\,\texttt{sweater\_\allowbreak{}middle} moves on the table $\geq$\,25\,cm \newline \textbullet\,the objects stand stacked, bottom to top: \texttt{sweater\_\allowbreak{}bottom} $\rightarrow$ \texttt{sweater\_\allowbreak{}top} \newline \textbullet\,\texttt{sweater\_\allowbreak{}bottom} moves on the table $\leq$\,3\,cm \\
\rowcolor{black!3}\vtop{\vskip-1.9ex\hbox{\includegraphics[width=1.35cm]{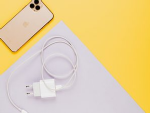}}} & \textbf{IC3} & ``Wind the cable into a neat coil and lay it next to the charger.'' & \textbullet\,\texttt{cable} gathers within a radius of $\leq$\,6\,cm (90\%) \newline \textbullet\,\texttt{cable} is let go at the end \newline \textbullet\,\texttt{phone} moves on the table $\leq$\,1\,cm \\
\vtop{\vskip-1.9ex\hbox{\includegraphics[width=1.35cm]{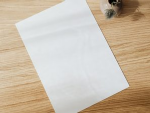}}} & \textbf{IC4} & ``Fold the sheet in half and press a sharp crease along the fold.'' & \textbullet\,\texttt{paper} has a sharp fold \newline \textbullet\,\texttt{paper} ends on 0.4--0.65 of its first footprint \newline \textbullet\,$\leq$\,5\% of \texttt{paper} wrinkles \newline \textbullet\,\texttt{vase} moves on the table $\leq$\,1\,cm \\
\rowcolor{black!3}\vtop{\vskip-1.9ex\hbox{\includegraphics[width=1.35cm]{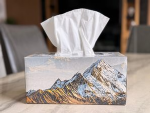}}} & \textbf{IC5} & ``Pull one tissue out of the box.'' & \textbullet\,only one \texttt{tissue\_\allowbreak{}*} ends $\geq$\,5 cm away from the \texttt{box} \newline \textbullet\,\texttt{tissue\_\allowbreak{}1} ends $\geq$\,5\,cm away from the \texttt{box} \newline \textbullet\,\texttt{box} moves on the table $\leq$\,2\,cm \newline \textbullet\,hands touch \texttt{tissue\_\allowbreak{}1} \\
\rowcolor{kflowing!14}\multicolumn{4}{@{\hspace{2pt}}l}{\rule{0pt}{9pt}\textcolor{kflowing}{\rule{5pt}{5pt}}\hspace{4pt}\textbf{Flowing}} \\
\vtop{\vskip-1.9ex\hbox{\includegraphics[width=1.35cm]{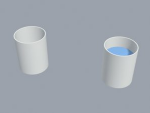}}} & \textbf{SF1} & ``Pour it into the empty cup.'' & \textbullet\,$\geq$\,80\% of \texttt{liquid} ends in or on the \texttt{cup\_\allowbreak{}empty} \newline \textbullet\,$\leq$\,5\% of \texttt{liquid} spills outside the \texttt{cup\_\allowbreak{}full} and \texttt{cup\_\allowbreak{}empty} \\
\rowcolor{black!3}\vtop{\vskip-1.9ex\hbox{\includegraphics[width=1.35cm]{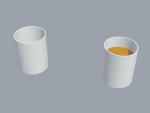}}} & \textbf{SF2} & ``Pour it into the empty cup.'' & \textbullet\,$\geq$\,60\% of \texttt{liquid} ends in or on the \texttt{cup\_\allowbreak{}empty} \newline \textbullet\,$\leq$\,5\% of \texttt{liquid} spills outside the \texttt{cup\_\allowbreak{}full} and \texttt{cup\_\allowbreak{}empty} \newline \textbullet\,\texttt{liquid} takes $\geq$\,2\,s to pour into the \texttt{cup\_\allowbreak{}empty} \\
\vtop{\vskip-1.9ex\hbox{\includegraphics[width=1.35cm]{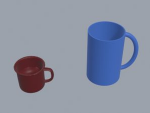}}} & \textbf{SF3} & ``Lift the pitcher by its handle and pour water into the mug.'' & \textbullet\,\texttt{mug} is filled to $\geq$\,30\% by the \texttt{water} \newline \textbullet\,$\leq$\,10\% of \texttt{water} spills outside the \texttt{pitcher} and \texttt{mug} \newline \textbullet\,\texttt{pitcher.\allowbreak{}handle} is held by the hand \newline \textbullet\,hands touch \texttt{pitcher.\allowbreak{}handle} \\
\rowcolor{black!3}\vtop{\vskip-1.9ex\hbox{\includegraphics[width=1.35cm]{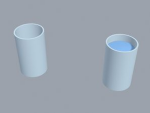}}} & \textbf{SF4} & ``Give me half a glass of water in the empty glass.'' & \textbullet\,\texttt{glass\_\allowbreak{}empty} is filled to 35--65\% by the \texttt{water} \newline \textbullet\,$\leq$\,5\% of \texttt{water} spills outside the \texttt{glass\_\allowbreak{}full} and \texttt{glass\_\allowbreak{}empty} \newline \textbullet\,hands stay 0.1\,cm clear of \texttt{water} \\
\vtop{\vskip-1.9ex\hbox{\includegraphics[width=1.35cm]{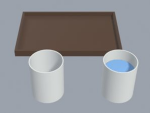}}} & \textbf{SF5} & ``Pour the water into the empty cup, then put both cups on the tray.'' & \textbullet\,$\geq$\,70\% of \texttt{water} ends in or on the \texttt{cup\_\allowbreak{}b} \newline \textbullet\,$\geq$\,90\% of \texttt{cup\_\allowbreak{}a} ends in or on the \texttt{tray} \newline \textbullet\,$\geq$\,90\% of \texttt{cup\_\allowbreak{}b} ends in or on the \texttt{tray} \newline \textbullet\,$\leq$\,5\% of \texttt{water} spills outside the \texttt{cup\_\allowbreak{}a} and \texttt{cup\_\allowbreak{}b} \\
\rowcolor{black!3}\vtop{\vskip-1.9ex\hbox{\includegraphics[width=1.35cm]{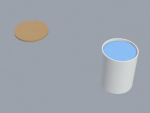}}} & \textbf{SF6} & ``Carry the cup to the coaster.'' & \textbullet\,$\geq$\,90\% of \texttt{cup} ends in the circle at (-12, 12) cm, r 6 cm \newline \textbullet\,$\leq$\,2\% of \texttt{water} spills outside the \texttt{cup} \newline \textbullet\,\texttt{cup} tilts $\leq$\,10$^\circ$ \newline \textbullet\,\texttt{cup} is let go at the end \newline \textbullet\,\texttt{cup} is at rest at the end ($\leq$\,5\,cm/s) \\
\vtop{\vskip-1.9ex\hbox{\includegraphics[width=1.35cm]{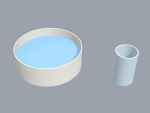}}} & \textbf{SF7} & ``Scoop up some water from the basin in your cupped hand and pour it into the glass.'' & \textbullet\,$\geq$\,0.3\% of \texttt{water} ends inside the \texttt{glass} \newline \textbullet\,$\leq$\,5\% of \texttt{water} spills outside the \texttt{basin} and \texttt{glass} \newline \textbullet\,\texttt{basin} moves $\leq$\,1\,cm \newline \textbullet\,\texttt{basin} tilts $\leq$\,5$^\circ$ \newline \textbullet\,hands stay 0.3\,cm clear of \texttt{glass} \\
\rowcolor{black!3}\vtop{\vskip-1.9ex\hbox{\includegraphics[width=1.35cm]{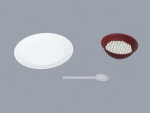}}} & \textbf{SF8} & ``Use the spoon to move some rice from the bowl to the plate.'' & \textbullet\,$\geq$\,3\% of \texttt{rice} ends in or on the \texttt{plate} \newline \textbullet\,hands touch \texttt{spoon.\allowbreak{}handle} \newline \textbullet\,hands stay 0.1\,cm clear of \texttt{rice} \\
\vtop{\vskip-1.9ex\hbox{\includegraphics[width=1.35cm]{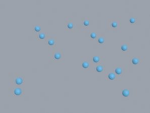}}} & \textbf{SF9} & ``Sweep the marbles together into one pile with your hand.'' & \textbullet\,\texttt{marbles} gathers within a radius of $\leq$\,6\,cm (90\%) \\
\rowcolor{black!3}\vtop{\vskip-1.9ex\hbox{\includegraphics[width=1.35cm]{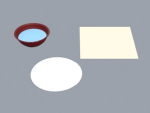}}} & \textbf{SF10} & ``Dip the cloth into the bowl of water, then lay it flat on the plate.'' & \textbullet\,$\geq$\,30\% of \texttt{cloth} is at some point inside the \texttt{bowl} \newline \textbullet\,$\geq$\,60\% of \texttt{cloth} ends in or on the \texttt{plate} \newline \textbullet\,\texttt{cloth} is let go at the end \newline \textbullet\,\texttt{bowl} tilts $\leq$\,15$^\circ$ \\
\vtop{\vskip-1.9ex\hbox{\textonlypanel}} & \textbf{TF1} & ``Share the juice equally among the three glasses.'' & \textbullet\,\texttt{glass\_\allowbreak{}1} is filled to 35--100\% by the \texttt{juice} \newline \textbullet\,\texttt{glass\_\allowbreak{}2} is filled to 35--100\% by the \texttt{juice} \newline \textbullet\,\texttt{glass\_\allowbreak{}3} is filled to 35--100\% by the \texttt{juice} \newline \textbullet\,the \texttt{glass\_\allowbreak{}1}, \texttt{glass\_\allowbreak{}2} and \texttt{glass\_\allowbreak{}3} hold \texttt{juice} within $\leq$\,15\% of each other \newline \textbullet\,$\leq$\,5\% of \texttt{juice} spills outside the \texttt{jug}, \texttt{glass\_\allowbreak{}1}, \texttt{glass\_\allowbreak{}2} and \texttt{glass\_\allowbreak{}3} \\
\rowcolor{black!3}\vtop{\vskip-1.9ex\hbox{\textonlypanel}} & \textbf{TF2} & ``Ladle some soup into the bowl without moving the pot.'' & \textbullet\,$\geq$\,5\% of \texttt{soup} ends in or on the \texttt{bowl} \newline \textbullet\,\texttt{pot} moves $\leq$\,1\,cm \newline \textbullet\,$\leq$\,5\% of \texttt{soup} spills outside the \texttt{pot}, \texttt{ladle} and \texttt{bowl} \newline \textbullet\,hands touch \texttt{ladle.\allowbreak{}handle} \\
\vtop{\vskip-1.9ex\hbox{\textonlypanel}} & \textbf{TF3} & ``Pour the sand through the funnel into the bottle.'' & \textbullet\,$\geq$\,70\% of \texttt{sand} ends inside the \texttt{bottle} \newline \textbullet\,$\leq$\,10\% of \texttt{sand} spills outside the \texttt{cup}, \texttt{funnel} and \texttt{bottle} \\
\rowcolor{black!3}\vtop{\vskip-1.9ex\hbox{\textonlypanel}} & \textbf{TF4} & ``Pour cereal into the bowl, then pour milk over it.'' & \textbullet\,$\geq$\,30\% of \texttt{cereal} ends in or on the \texttt{bowl} \newline \textbullet\,$\geq$\,20\% of \texttt{milk} ends in or on the \texttt{bowl} \newline \textbullet\,$\leq$\,10\% of \texttt{cereal} spills outside the \texttt{box} and \texttt{bowl} \newline \textbullet\,$\leq$\,10\% of \texttt{milk} spills outside the \texttt{carton} and \texttt{bowl} \\
\vtop{\vskip-1.9ex\hbox{\textonlypanel}} & \textbf{TF5} & ``Drizzle some honey over the toast with the dipper.'' & \textbullet\,$\geq$\,2\% of \texttt{honey} ends in or on the \texttt{toast} \newline \textbullet\,\texttt{dipper.\allowbreak{}handle} is held by the hand \newline \textbullet\,$\leq$\,5\% of \texttt{honey} spills outside the \texttt{jar}, \texttt{dipper}, \texttt{toast} and \texttt{plate} \newline \textbullet\,hands stay 0.5\,cm clear of \texttt{honey} \\
\rowcolor{black!3}\vtop{\vskip-1.9ex\hbox{\includegraphics[width=1.35cm]{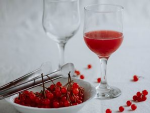}}} & \textbf{IF1} & ``Pour half of the wine into the empty glass.'' & \textbullet\,35--65\% of \texttt{wine} ends in or on the \texttt{glass\_\allowbreak{}empty} \newline \textbullet\,$\leq$\,5\% of \texttt{wine} spills outside the \texttt{glass\_\allowbreak{}full} and \texttt{glass\_\allowbreak{}empty} \newline \textbullet\,\texttt{berry\_\allowbreak{}bowl} moves on the table $\leq$\,1\,cm \newline \textbullet\,hands stay 0.1\,cm clear of \texttt{wine} \\
\vtop{\vskip-1.9ex\hbox{\includegraphics[width=1.35cm]{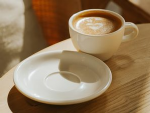}}} & \textbf{IF2} & ``Set the cup on its saucer without spilling the coffee.'' & \textbullet\,$\geq$\,90\% of \texttt{cup} ends in or on the \texttt{saucer} \newline \textbullet\,$\leq$\,2\% of \texttt{coffee} spills outside the \texttt{cup} \newline \textbullet\,\texttt{cup} tilts $\leq$\,12$^\circ$ \newline \textbullet\,\texttt{cup} is let go at the end \newline \textbullet\,\texttt{cup} is at rest at the end ($\leq$\,5\,cm/s) \newline \textbullet\,hands stay 0.1\,cm clear of \texttt{coffee} \newline \textbullet\,hands touch \texttt{cup.\allowbreak{}handle} \\
\rowcolor{black!3}\vtop{\vskip-1.9ex\hbox{\includegraphics[width=1.35cm]{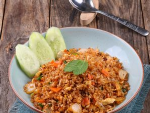}}} & \textbf{IF3} & ``Scoop up a heaping spoonful of rice and hold it up without dropping any.'' & \textbullet\,$\geq$\,2\% of \texttt{rice} ends in or on the \texttt{spoon} \newline \textbullet\,\texttt{spoon} rises $\geq$\,8\,cm \newline \textbullet\,$\leq$\,1\% of \texttt{rice} spills outside the \texttt{bowl} and \texttt{spoon} \newline \textbullet\,hands stay 0.1\,cm clear of \texttt{rice} \newline \textbullet\,hands touch \texttt{spoon.\allowbreak{}handle} \\
\vtop{\vskip-1.9ex\hbox{\includegraphics[width=1.35cm]{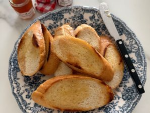}}} & \textbf{IF4} & ``Spread some apricot jam on the front slice of toast with the knife.'' & \textbullet\,\texttt{jam} covers $\geq$\,25\% of the \texttt{toast} \newline \textbullet\,\texttt{knife.\allowbreak{}handle} is held by the hand \newline \textbullet\,$\leq$\,5\% of \texttt{jam} spills outside the \texttt{apricot\_\allowbreak{}jar}, \texttt{knife}, \texttt{toast} and \texttt{plate} \newline \textbullet\,hands stay 0.2\,cm clear of \texttt{jam} \newline \textbullet\,hands touch \texttt{knife.\allowbreak{}handle} \\
\rowcolor{black!3}\vtop{\vskip-1.9ex\hbox{\includegraphics[width=1.35cm]{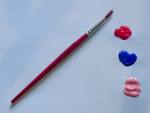}}} & \textbf{IF5} & ``Mix the red and blue paint together with the brush until it is purple, without touching the pink.'' & \textbullet\,$\geq$\,30\% of the surface of \texttt{red\_\allowbreak{}paint} and the \texttt{blue\_\allowbreak{}paint} ends mixed \newline \textbullet\,\texttt{brush.\allowbreak{}handle} is held by the hand \newline \textbullet\,\texttt{pink\_\allowbreak{}paint} moves on the table $\leq$\,0.5\,cm \newline \textbullet\,\texttt{pink\_\allowbreak{}paint} ends $\geq$\,0.3\,cm away from the \texttt{red\_\allowbreak{}paint} \newline \textbullet\,\texttt{pink\_\allowbreak{}paint} ends $\geq$\,0.3\,cm away from the \texttt{blue\_\allowbreak{}paint} \newline \textbullet\,hands stay 0.2\,cm clear of \texttt{red\_\allowbreak{}paint} \newline \textbullet\,hands stay 0.2\,cm clear of \texttt{blue\_\allowbreak{}paint} \newline \textbullet\,hands stay 0.2\,cm clear of \texttt{pink\_\allowbreak{}paint} \newline \textbullet\,hands touch \texttt{brush.\allowbreak{}handle} \\
\end{longtable}
}

\subsection{Success Condition Metrics}
\label{app:cat:metrics}
Each success condition in \Cref{tab:cat_tasks} compares a metric of a named object or part with a threshold.
\Cref{tab:cat_metrics} defines these metrics with their units and the tasks that use them.
Fractions are taken over surface vertices or volume samples.

{\scriptsize
\setlength{\tabcolsep}{3pt}\renewcommand{\arraystretch}{1.2}
\begin{longtable}{@{}>{\raggedright\arraybackslash}p{\dimexpr 0.18\linewidth-2\tabcolsep\relax}>{\raggedright\arraybackslash}p{\dimexpr 0.59\linewidth-2\tabcolsep\relax}>{\centering\arraybackslash}p{\dimexpr 0.07\linewidth-2\tabcolsep\relax}>{\raggedright\arraybackslash}p{\dimexpr 0.16\linewidth-2\tabcolsep\relax}@{}}
  \caption{\small Metrics of the success conditions by group, each with its definition, unit and the tasks that use it.}
  \label{tab:cat_metrics}\\
  \toprule
  \textbf{Metric} & \textbf{Definition} & \textbf{Unit} & \textbf{Tasks} \\
  \midrule
  \endfirsthead
  \multicolumn{4}{@{}l}{\small\emph{\Cref{tab:cat_metrics}, continued}} \\
  \toprule
  \textbf{Metric} & \textbf{Definition} & \textbf{Unit} & \textbf{Tasks} \\
  \midrule
  \endhead
  \bottomrule
  \endfoot
  \rowcolor{black!7}\multicolumn{4}{@{\hspace{2pt}}l}{\rule{0pt}{9pt}\textbf{Placement and Motion}} \\
Rise & Maximum rise of the centroid & cm & IF3, SR5, SR8 \\
\rowcolor{black!3}Displacement & Final displacement of the centroid along an axis, in the table plane or in 3D & cm & 21 tasks \\
Maximum tilt & Maximum tilt of the object's up axis during the motion & $^\circ$ & IF2, IR2, SF6, SF7, SF10 \\
\rowcolor{black!3}Final orientation & Final angle between an object axis and a target world direction & $^\circ$ & 6 tasks \\
Transferred fraction & Fraction of the object in or on the receiver at the end & \% & 20 tasks \\
\rowcolor{black!3}Fraction inside & Fraction of the object inside the receiver below its rim, at the end or at its peak & \% & 6 tasks \\
Fraction in a region & Fraction of the object inside a target circle on the table at the end & \% & SF6 \\
\rowcolor{black!3}Final gap & Final gap between the object and another object & cm & IC5, IF5, IR5, SC4, TR4 \\
Lowest point & Final height of the object's lowest point & cm & IR5, SC8, SD8, SR7, TC2 \\
\rowcolor{black!3}Rest & Maximum centroid speed over the last 0.5\,s & cm/s & 12 tasks \\
Release & No hand contact over the last 0.3\,s & -- & 27 tasks \\
\rowcolor{black!3}Stacking & Objects stacked in the given order, each covering at least half of the one below & -- & 6 tasks \\
Untouched travel & Longest distance traveled without hand contact after the first touch (\eg, a throw) & cm & SR7, TC5 \\
\rowcolor{black!3}Flight height & Maximum rise of the centroid while the object touches nothing & cm & SR6 \\
Strike & Contact with the target after the object leaves the hand & -- & SR7 \\
\rowcolor{black!3}Hops & Number of times the object leaves the table by over 2\,cm and lands again & -- & SD5 \\
\rowcolor{black!7}\multicolumn{4}{@{\hspace{2pt}}l}{\rule{0pt}{9pt}\textbf{Parts and Hands}} \\
Part rotation & Rotation of a part relative to its parent (\eg, a lid), at the end or at its maximum & $^\circ$ & SR8, SR9, SR10, TR5 \\
\rowcolor{black!3}Finger contact & Joint contact of the given fingers for at least 0.2\,s & -- & 12 tasks \\
\rowcolor{black!7}\multicolumn{4}{@{\hspace{2pt}}l}{\rule{0pt}{9pt}\textbf{Shape}} \\
Height ratio & Final over initial height & -- & ID1, SD3, TD1 \\
\rowcolor{black!3}Volume change & Relative change of the closed volume, $|V/V_0-1|$ & \% & ID4, SD4 \\
Compression & Maximum relative reduction of the object's width during the motion & \% & 6 tasks \\
\rowcolor{black!3}Recovery & Final over initial shortest width, which tests elastic recovery & -- & ID5, SD2, SD6 \\
Length ratio & Final over initial length & -- & ID4 \\
\rowcolor{black!3}Dent depth & Maximum depth of the surface pressed in by a hand & cm & SD2 \\
Kept dent & Depth of the dent that remains after the hand leaves & cm & ID1, SD1, SD3 \\
\rowcolor{black!3}Hollow area & Area of the largest hollow in the top that would hold water 1\,cm deep & cm$^2$ & SD3 \\
Elongation & Final ratio of the longest to the middle extent & -- & ID4, SD4 \\
\rowcolor{black!3}Stretch & Maximum over initial length & -- & TD3 \\
Pieces & Number of final pieces that each hold a given fraction of the object & -- & ID2, SC7, SD7, TD5 \\
\rowcolor{black!7}\multicolumn{4}{@{\hspace{2pt}}l}{\rule{0pt}{9pt}\textbf{Cloth, Paper and Rope}} \\
Footprint ratio & Final over initial footprint area, which measures folding or bunching & -- & 8 tasks \\
\rowcolor{black!3}Crease & Fraction of the main fold bent by over 120$^\circ$, which tests a sharp fold & \% & IC4, SC5 \\
Crumpling & Fraction of the sheet away from one fold that bends by over 60$^\circ$ & \% & SC6 \\
\rowcolor{black!3}Wrinkling & Fraction of the sheet away from one fold that bends by over 30$^\circ$ & \% & IC4, SC5 \\
Flatness along a line & Fraction of a ribbon lying flat along one line & \% & SC2 \\
\rowcolor{black!3}Lifted fraction & Fraction of the object more than 1\,cm above the table at the end & \% & IC1 \\
Top coverage & Fraction of the target's top covered by the object & \% & IF4 \\
\rowcolor{black!3}Wrapping coverage & Fraction of the target's top and sides covered by the object & \% & TC3 \\
Separated pieces & Number of named pieces that end at least a given distance from another object & -- & IC5, SC4 \\
\rowcolor{black!3}Wiped coverage & Fraction of the target that the object wipes over & \% & SC10 \\
Wrap angle & Angle that a cord's centerline sweeps around the receiver's long axis & $^\circ$ & SC9, TC2 \\
\rowcolor{black!7}\multicolumn{4}{@{\hspace{2pt}}l}{\rule{0pt}{9pt}\textbf{Grains, Liquids and Paint}} \\
Spill & Fraction of the material spilled outside every container & \% & 17 tasks \\
\rowcolor{black!3}Fill level & Liquid height in the receiver over its inner height & \% & SF3, SF4, TF1 \\
Fill spread & Range of the fill levels across the containers relative to the highest & \% & TF1 \\
\rowcolor{black!3}Gathering radius & Radius of the circle that holds 90\,\% of the material & cm & IC3, SC2, SF9, TC4 \\
Pouring time & Time for the fraction in the receiver to rise from 10\,\% to 90\,\% & s & SF2 \\
\rowcolor{black!3}Mixing & Fraction of the paint surface with a color between the two paints & \% & IF5 \\
\end{longtable}
}

\subsection{Physical Checks}
\label{app:cat:checks}
Whereas each task has its own success conditions, the physical checks in \Cref{tab:cat_checks} apply to every task.
They cover the hands in matter, carrying and object dynamics, and they determine Phys and SR as described in \Cref{app:metrics}.

\vspace{0.3em}
\noindent\begin{minipage}{\linewidth}
\captionof{table}{Physical checks shared by all tasks. Credit falls linearly from full to zero credit, and a solved task meets every threshold.}
\label{tab:cat_checks}
\scriptsize\setlength{\tabcolsep}{3pt}\renewcommand{\arraystretch}{1.2}
\begin{tabular}{@{}>{\raggedright\arraybackslash}p{\dimexpr 0.18\linewidth-2\tabcolsep\relax}>{\raggedright\arraybackslash}p{\dimexpr 0.5\linewidth-2\tabcolsep\relax}>{\centering\arraybackslash}p{\dimexpr 0.06\linewidth-2\tabcolsep\relax}>{\centering\arraybackslash}p{\dimexpr 0.09\linewidth-2\tabcolsep\relax}>{\centering\arraybackslash}p{\dimexpr 0.09\linewidth-2\tabcolsep\relax}>{\centering\arraybackslash}p{\dimexpr 0.08\linewidth-2\tabcolsep\relax}@{}}
\toprule
  \textbf{Check} & \textbf{Definition} & \textbf{Unit} & \textbf{Full Credit} & \textbf{Zero Credit} & \textbf{Solved} \\
\midrule
\rowcolor{black!7}\multicolumn{6}{@{\hspace{2pt}}l}{\rule{0pt}{9pt}\textbf{Hands in Matter}} \\
Hand-solid penetration & Deepest mutual penetration of a hand and a solid & mm & 2 & 10 & $\leq$\,5 \\
\rowcolor{black!3}Points inside a hand & Fraction of cloth, rope, grain or liquid points inside a hand & \% & 0.5 & 10 & $\leq$\,5 \\
Pins away from hands & Cloth or rope pinned away from the hands and dragged over 2\,cm or held aloft over 0.5\,s & -- & No & Yes & No \\
\rowcolor{black!7}\multicolumn{6}{@{\hspace{2pt}}l}{\rule{0pt}{9pt}\textbf{Carrying}} \\
Infeasible grip & Fraction of carried frames whose contacts cannot support the weight & \% & 25 & 75 & $\leq$\,50 \\
\rowcolor{black!3}Slip in hand & Slip of the skin over a carried object & mm & 15 & 40 & $\leq$\,30 \\
Unsupported tipping & Fraction of resting frames with the center of mass over 5\,mm outside the support & \% & 10 & 30 & $\leq$\,20 \\
\rowcolor{black!7}\multicolumn{6}{@{\hspace{2pt}}l}{\rule{0pt}{9pt}\textbf{Object Dynamics}} \\
Unsupported object motion & Fraction of frames with an untouched object rising or resting on nothing & \% & 0 & 10 & $\leq$\,10 \\
\rowcolor{black!3}Unsupported loose matter & Fraction of frames with over 15\,\% of the grains or liquid resting on nothing & \% & 0 & 10 & $\leq$\,10 \\
Sinking into a support & Deepest penetration of an object into the table or another solid & mm & 4 & 15 & $\leq$\,10 \\
\rowcolor{black!3}Sheet or cord stretch & 95th-percentile edge stretch of cloth or rope in the worst frame & -- & 1.15 & 1.6 & $\leq$\,1.5 \\
Rigid shape residual & Largest deviation of a rigid object from a rigid motion & mm & 2 & 10 & $\leq$\,5 \\
\rowcolor{black!3}Unforced deformation & Shape change of a deformable solid while untouched & mm & 5 & 15 & $\leq$\,10 \\
\bottomrule
\end{tabular}
\end{minipage}